\documentclass[sigconf]{acmart}
\usepackage{multirow}
\usepackage{subfigure}
\usepackage{bbm}
\usepackage{bm}
\usepackage{algorithmicx,algorithm}
\usepackage{enumitem}
\usepackage{xcolor}

\AtBeginDocument{%
  }

\setcopyright{acmcopyright}
\copyrightyear{2023}
\acmYear{2023}
\acmDOI{XXXXXXX.XXXXXXX}

\acmConference[SIGMOD'23]{the 2023 International Conference on Management of Data, June 18--23, 2023, Seattle, WA, USA.}
\acmPrice{15.00}
\acmISBN{978-1-4503-XXXX-X/18/06}

\begin{document}


\title{Joint Neural Architecture and Hyperparameter Search for Correlated Time Series Forecasting}




\author{Xinle Wu, Dalin Zhang$^{\ast}$, Miao Zhang, Chenjuan Guo, Bin Yang$^{\ast}$, Christian S. Jensen\\
Department of Computer Science, Aalborg University, Denmark
}
\email{{xinlewu, dalinz, miaoz, cguo, byang, csj}@cs.aau.dk}\thanks{*corresponding authors: D.~Zhang (dalinz@cs.aau.dk) and B.~Yang (byang@cs.aau.dk)}

\begin{abstract}

Sensors in cyber-physical systems often capture interconnected processes and thus emit correlated time series (CTS), the forecasting of which enables important applications. The key to successful CTS forecasting is to uncover the temporal dynamics of time series and the spatial correlations among time series. Deep learning-based solutions exhibit impressive performance at discerning these aspects. In particular, automated CTS forecasting, where the design of an optimal deep learning architecture is automated, enables forecasting accuracy that surpasses what has been achieved by manual approaches.  However, automated CTS solutions remain in their infancy and are only able to find optimal architectures for predefined hyperparameters and scale poorly to large-scale CTS. To overcome these limitations, we propose SEARCH, a joint, scalable framework, to automatically devise effective CTS forecasting models. Specifically, we encode each candidate architecture and accompanying hyperparameters into a joint graph representation. We introduce an efficient Architecture-Hyperparameter Comparator (AHC) to rank all architecture-hyperparameter pairs, and we then further evaluate the top-ranked pairs to select a final result. Extensive experiments on six benchmark datasets demonstrate that SEARCH not only eliminates manual efforts but also is capable of better performance than manually designed and existing automatically designed CTS models. In addition, it shows excellent scalability to large CTS.

\end{abstract}





\maketitle

\section{Introduction}

Many systems, including societal infrastructures such as transportation systems, electricity grids, and sewage systems~\cite{rajkumar2010cyber}, include cyber-physical components that encompass multiple sensors that each emit a time series, resulting in multiple time series that are often correlated.
For example, inductive-loop detectors in a vehicular transportation system measure the time-varying traffic volume at different road locations, and measurements along the same or nearby roads often correlate.
The forecasting of future values from correlated time series often has important applications\cite{DBLP:conf/ijcai/WuPLJZ19}. For example, accurate forecasting of traffic volumes can facilitate the prediction of congestion and near-future travel times, in turn enabling, e.g., more effective vehicle routing. 

The key to successful correlated time series forecasting is the ability to capture both the temporal dependencies among historical values of each time series and the spatial correlations across different time series. Leveraging the powerful feature extraction capabilities of deep learning models, different neural architectures, called ST-blocks, have been proposed to capture spatio-temporal (ST) dependencies to enable accurate forecasting. 
%
Traditionally, human experts have designed ST-blocks and have chosen accompanying hyperparameter settings manually. However, this is a resource-intensive endeavor for which human expertise is ill-suited.

\begin{figure*}[!htbp]
\center
\subfigure[Existing Supernet-based Framework]{
\begin{minipage}[c]{0.33\linewidth} 
\centering
\includegraphics[width=\linewidth]{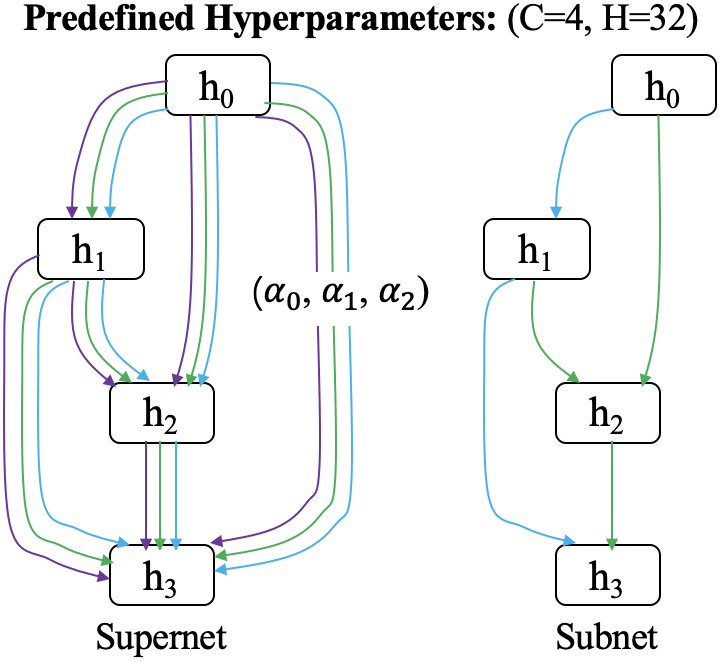}
\label{fig: exist}
\end{minipage}
}
\subfigure[The SEARCH Framework]{
\begin{minipage}[c]{0.63\linewidth} 
\centering
\includegraphics[width=\linewidth]{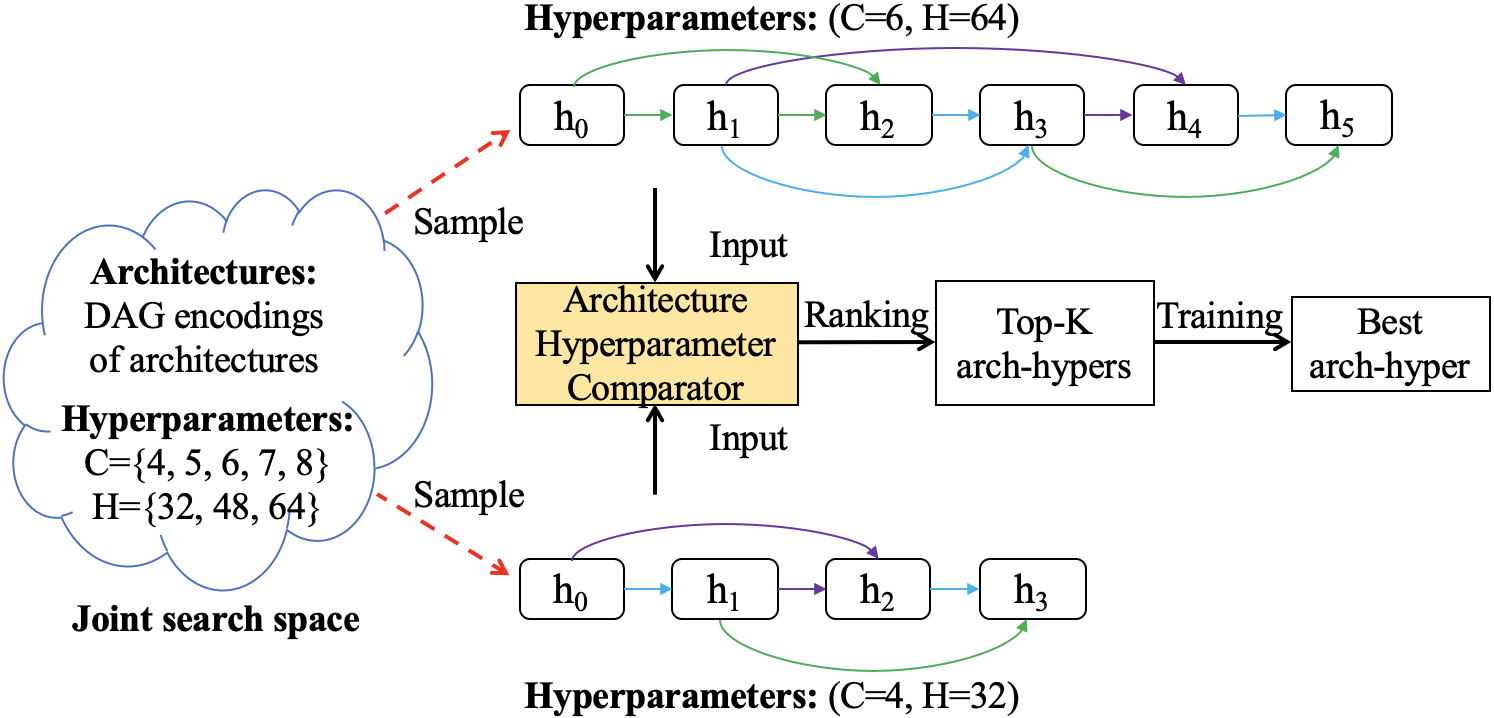}
\label{fig: framework}
\end{minipage}
}
\caption{
Comparsion between exisitng supernet-based framework vs. the proposed SEARCH framework. 
Hyperparameter $C$ indicates the number of nodes in an ST-block, and $H$ is the size of hidden representations. Edges of different colors represent different operators.
}
\label{fig: compare}
\end{figure*}

A more recent approach is to automate the design of effective ST-blocks~\cite{pan2019urban,wu2022autocts}.
Figure~\ref{fig: exist} outlines a typical  automated framework. A search space, represented as a supernet, contains a massive number of possible ST-blocks (Figure~\ref{fig: exist} left), any subnet of which is a candidate ST-block (Figure~\ref{fig: exist} right). Nodes in the supernet and subnets represent latent representations, and the directed edges between them represent different operators (e.g., convolution, graph convolution, and Transformer). In a supernet, the transformation from node $h_i$ to node $h_j$ is a weighted sum of all candidate operators, while in a subnet only one operator between each node pair is kept. The goal is to learn the operator-associated weights $\{\alpha_i\}$, upon which an optimal subnet is obtained by picking the operator with the highest weight between every two nodes.

Although existing automated CTS forecasting methods achieve design automation and are capable of better performance than manually designed models, they still suffer from three main limitations.

\noindent \textbf{(1) Lack of support for joint search for architectures and hyperparameters.}
When training supernets, existing automated CTS forecasting methods rely on predefined hyperparameters, including architectural hyperparameters (e.g., the number of latent representations, i.e., nodes, in an ST-block and the size of a latent representation) 
and training hyperparameters (e.g., the dropout rate). Depending on the hyperparameter settings, the same architecture can yield markedly different performance. In spite of this, existing solutions rely on an expert to select proper hyperparameters settings, which may well lead to the choice of a suboptimal architecture and which renders the framework ``semi-automated.''
Alternatively, it is possible to combine a hyperparameter optimization method (e.g., grid search or Bayesian optimization) sequentially with an existing automated architecture search method \cite{pan2019urban,wu2022autocts} to achieve a two-step, automated approach. However, the running time would be excessive as existing automated architecture search needs to be performed each time a hyperparameter set is sampled. Rather, an efficient, joint architecture and hyperparameter search scheme is called for.

\noindent \textbf{(2) Poor scalability.}
Existing automated CTS forecasting methods usually suffer from poor scalability, as an entire supernet must reside in memory during training, which may cause memory overflow in large-scale CTS settings~\cite{pan2019urban,wu2022autocts}.
Specifically, the memory cost of the neural operators that compose a supernet increases rapidly with the number of time series $N$ and the number of historical timestamps $P$ in time series. For example, the memory cost of commonly used one-dimensional convolutional neural networks (1D CNNs, T-operators) and Graph Convolution Networks (GCNs, S-operators) grows linearly with $N$ and $P$, and the memory cost of Transformers (S/T-operators) grows quadratically with $P$ and $N$. Since there are $O$ operators between each node pair in a supernet, the memory cost of a supernet is approximately $O$ times that of a subnet, i.e., an actual CTS forecasting model.
Taking Figure~\ref{fig: exist} as an example, the supernet has three types of candidate operators between each pair of nodes, i.e., 1D CNN, GCN, and Transformer (represented by different colored lines). The memory cost of the supernet is thus about three times that of one of its subnets.
When increasing $N$ or $P$, the memory usage of a supernet goes overflow more than a subnet, which limits the scalability of neural architecture search.

\noindent \textbf{(3) One-time use.}
Existing automated CTS forecasting methods train a supernet for each specific dataset from scratch, which is costly~\cite{pan2019urban,wu2022autocts}. 
%
However, given the possible similarities between different CTS forecasting datasets, reusing knowledge learned from previous search on relevant datasets may significantly reduce the search time on a new dataset. Therefore, developing a transferable automated framework holds potential to increase search efficiency.
%


%
We propose SEARCH, a Scalable and Efficient joint ARChitecture and Hyperparameter search framework to address the above limitations.
First, we design a joint search space that contains a wide variety of Architecture-Hyperparameter (arch-hyper) combinations, and then aim at finding the optimal arch-hyper in this search space, thus addressing the first limitation. 
%
For example, with existing supernet based methods, the number $C$ of nodes in an ST-block must be set before searching. The example in Figure~\ref{fig: exist} can only search for ST-blocks with $C=4$ nodes. 
In contrast, our joint search space considers multiple values for $C$, e.g., \{4, 5, 6, 7, 8\} in Figure~\ref{fig: framework}, and allows searching for ST-blocks with different numbers of nodes. 
%
%

Second, to avoid using a supernet that consumes extensive memory and limits scalability, we propose a novel Architecture Hyperparamter Comparator (AHC) to rank candidate arch-hypers from the joint search space. 
%
Given encodings of two candidate arch-hypers, the AHC 
estimates a binary value, indicating which arch-hyper has the best accuracy. 
Thus, AHC is able to estimate a ranking of candidate arch-hypers. We then train the top-K ranked arch-hypers and select the arch-hyper that gives the highest forecasting accuracy. 
Because the AHC is implemented with a lightweight graph neural network 
and the input to the AHC is the encodings of arch-hypers instead of CTS data, its performance is independent of $P$ and $N$, making it more scalable than existing supernet-based frameworks.

Third, instead of learning a new AHC on each unseen dataset from scratch, we propose to transfer the AHC trained on one dataset to unseen datasets, thereby improving the training efficiency on the unseen datasets, thus addressing the third limitation.

%
To the best of our knowledge, this is the first study that enables joint search for architectures and corresponding hyperparameter settings for correlated time series forecasting.
Specifically, we make the following contributions:
\begin{itemize}
    \item [(1)]
    We propose a novel search space for correlated time series forecasting to facilitate joint search for architectures and hyperparameter settings.
    \item [(2)]
   A memory-efficient Architecture-Hyperparameter Comparator (AHC) is proposed to rank arch-hyper candidates that encompasses an easy-to-obtain proxy metric to generate pseudo-labels to train an AHC in a denoising manner, thereby improving search efficiency.
    \item [(3)]
    We propose a method 
    that is able to quickly adapt a trained AHC to a new dataset, 
    thus significantly improving the AHC training efficiency on new datasets. 
    \item [(4)]
    Extensive experiments on six benchmark datasets show that SEARCH is able to efficiently find better-performing CTS forecasting models compared to state-of-the-art manual and automatic methods.
\end{itemize}


\section{Preliminaries}
\label{Sec: Prelim}

\subsection{Problem Settings}

\noindent\textbf{Correlated Time Series (CTS).}
We denote 
$N$ correlated time series (CTS) by $\bm{\mathcal{X}} \in \mathbb{R}^{N\times T\times F}$, where each time series contains $T$ timestamps and has an $F$-dimension feature vector at each timestamp.
The 
feature vectors in the $i$-th time series $\bm{X}^{(i)} \in \mathbb{R}^{T\times F}$ $\subset \bm{\mathcal{X}}$, $1\leq i\leq N$, are correlated with previous feature vectors in the time series as well as feature {vectors} in other time series.
Therefore, it is natural to model a CTS as a graph $G=(V,E,A)$, where vertex set $V$ represents the set of time series, edge set $E$ represents correlation relationships between time series, and adjacency matrix 
$A \in \mathbb{R}^{N\times N}$ captures the strengths of the relationships between time series. $A$ is usually predefined based on the distances of the sensors that generate the time series, or learned adaptively.

\noindent\textbf{Correlated Time Series Forecasting.}
We consider multi-step and single-step correlated time series forecasting{, both of which have important applications in the real world~\cite{shih2019temporal,lai2018modeling,wu2020connecting,bai2020adaptive}}. 
Given the feature vectors of the past $P$ time steps of $\bm{\mathcal{X}}$, the goal of multi-step CTS forecasting is to predict the feature vectors of the future $Q$ time step, with $Q>1$; and the goal of single-step CTS forecasting is to predict the vector at the $Q$-th future time step, where $Q\geq 1$.
Formally, we define the multi-step CTS forecasting as follows:
\begin{align}
{(\bm{\hat{X}}_{t+P+1}, \bm{\hat{X}}_{t+P+2}, ...,\bm{\hat{X}}_{t+P+Q})}=\mathcal{F}(\bm{X}_{t+1}, \bm{X}_{t+2}, \ldots, \bm{X}_{t+P};G),
\label{form: multi}
\end{align}
and we define the single-step CTS forecasting as follows:
\begin{align}
\bm{\hat{X}}_{t+P+Q}=\mathcal{F}(\bm{X}_{t+1}, \bm{X}_{t+2}, \ldots, \bm{X}_{t+P};G),
\label{form: single}
\end{align}
where $\bm{X}_{t} \in \mathbb{R}^{N\times F}$ denotes the feature vectors of all time series at timestamp $t$; 
while $\bm{\hat{X}}$ represents the predicted feature vectors, and $\mathcal{F}$ is a CTS forecasting model.

\noindent\textbf{Problem Definition.}
The goal is to automatically build an optimal ST-block $\mathcal{F}^{\ast}$
from a predefined combined architecture-hyperparameter search space $\mathcal{S}$ that minimizes the forecasting error on a validation dataset $\mathcal{D}_{val}$.
Mathematically, the objective function can be stated as the following equation:
\begin{align}
\mathcal{F}^{\ast}=
\mathit{argmin}_{\mathcal{F} \in \mathcal{S}}~ \mathit{ErrorMetric}(\mathcal{F},\mathcal{D}_{val})
\label{form: 2.1}
\end{align}

\subsection{Neural Forecasting Models}
\label{ssec: manually}
\begin{figure}[htb]
  \centering
  \includegraphics[width=\linewidth]{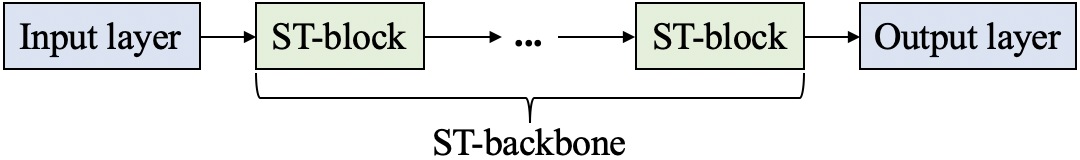}
  \caption{An example CTS forecasting model. 
  }
  \label{fig: arch}
\end{figure}

As summarized in Figure~\ref{fig: arch}, the common framework of manually designed neural CTS forecasting models has three components: an input module, an ST-backbone, and an output module. The input and output modules usually consist of one or two fully-connected layers that encode the input time series and decode extracted spatiotemporal features to forecasting values, respectively. 

The ST-backbone is the core component of a CTS forecasting model. It consists of $B$ ST-blocks.
The $B$ ST-blocks can be connected using different topologies, with sequential stacking being a simple yet effective topology and that we use as an example in Figure~\ref{fig: arch}.
An ST-block captures the spatial correlations between time series and the temporal dependencies in individual time series. There are thus two categories of operators in an ST-block, S-operators (e.g., GCNs) and T-operators (e.g., Transformer), for extracting spatial and temporal features, respectively. The specific types of S/T-operators and their connections are critical to the success of a CTS forecasting model.

\subsection{Existing Automated Methods}
Since the input and output modules of neural CTS forecasting models are simple, typically consisting of one or two fully-connected layers, and only involve a few design {choices}, such as the choice of embedding dimension, existing automated CTS forecasting frameworks ~\cite{wu2022autocts,pan2019urban} focus on the design of ST-blocks.
Specifically, automated frameworks typically start by designing a search space that encompasses a wide variety of architectures for ST-blocks.
The search space is represented by a directed acyclic graph (DAG) (Figure~\ref{fig: exist} left), dubbed a supernet, with $C$ nodes and a number of edges. Each node $h_i$, $0\leq i\leq C-1$, denotes a latent representation. Each node pair $(h_i, h_j)$, has $|\mathcal{O}|$ directed edges from $h_i$ to $h_j$, $i<j$, corresponding to $|\mathcal{O}|$ candidate S/T-operators, where $\mathcal{O}$ is a predefined candidate operator set consisting of S/T-operators, such as CNN, GCN, and Transformer.

The goal of existing automated frameworks is to select operators and connections that minimize validation errors to obtain ST-blocks and a consequent CTS forecasting model.
To achieve this, a vector $\alpha^{(i,j)}\in \mathbb{R}^{|\mathcal{O}|}$ is introduced to weigh the edges between each node pair $(h_i, h_j)$. These vectors then reflect the importance of the edges and are to be learned during training.
Then the transformation from node $h_i$ to node $h_j$ is formulated as a weighted sum of all edges, i.e., operators:
\begin{align}
f^{(i,j)}=\sum\limits_{o\in \mathcal{O}}\frac{exp(\alpha_o^{(i,j)})}{\sum_{o'\in \mathcal{O}}exp(\alpha_{o'}^{(i,j)})}o(h_i), 
\label{2.3.1}
\end{align}
where $\alpha_o^{(i,j)}$ is the weight of operator $o\in\mathcal{O}$, and $o(\cdot)$ represents the transform function of operator $o$.
Then, the latent representation of node $h_j$ is obtained by summing all the transformations from its predecessor nodes:
\begin{align}
h_j=\sum\limits_{i<j}f^{(i,j)}
\label{2.3.2}
\end{align}
This way, a supernet can be trained on the target CTS forecasting dataset using gradient descent to learn both the neural operator parameters and architecture parameter $\alpha$. After training, an optimal ST-block is derived by removing the unimportant edges from the supernet, retaining only one edge between each node pair and at most two incoming edges for each node (Figure~\ref{fig: exist} right).

For existing automated frameworks, hyperparameters such as the number $C$ of nodes in an ST-block need to be predefined (e.g., C is set to 4 in Figure~\ref{fig: exist}). In other words, existing frameworks do not support jointly searching for architectures and hyperparameters. In addition, existing automated frameworks consume substantial memory since very large supernets must reside in memory during training. Furthermore, existing automated frameworks start from scratch for each new dataset, which is inefficient.

\section{{scalable} and Efficient Joint Search}
\label{sec: method}
Our framework enables search for an optimal ST-block, i.e., an optimal combination of a neural architecture and a set of accompanying hyperparameters. Figure~\ref{fig: framework} offers an overview of the proposed automated CTS forecasting framework.
In order to support joint search, we first design a {\bf joint search space} (Section~\ref{ssec: space}) containing various candidate [architecture, hyperparameters] pairs, each of which is denoted as an {\it arch-hyper}. 

We then propose a novel search framework that leverages an {\bf Architecture-Hyperparameter Comparator (AHC)} (Section~\ref{ssec: AHC}) to achieve a predicted accuracy based ranking of all arch-hypers in the joint search space.
The AHC is implemented as a neural network that takes the encodings of two candidate arch-hypers as input and produces a binary label indicating which input arch-hyper has the higher accuracy. 
To this end, training the AHC requires a large number of labeled samples of the form ($ah_1$, $ah_2$, $y$),
where $ah_1$ and $ah_2$ are arch-hypers and $y$ is a binary label. 
Intuitively, obtaining the labels requires completely training the two input arch-hypers, which is computationally expensive. 
To reduce cost, we propose an easy-to-get {\bf proxy metric} (Section~\ref{ssec: proxy}) to generate pseudo-labels and further train the AHC in a noise reduction manner to reduce the negative impact of the pseudo-labels. 
Once an arch-hyper ranking is obtained, we pick the top-$K$ arch-hypers for full training, and we finally select the optimal arch-hyper with the highest validation accuracy.
Furthermore, we propose a simple yet effective {\bf transfer} method (Section~\ref{ssec: transfer}) to avoid learning an AHC from scratch on each new dataset, which is achieved by transferring an AHC that is trained on one dataset to a new dataset. Thus, only few AHC training samples from the new dataset are required to finetune the pretrained AHC, which significantly improves the search efficiency without compromising the accuracy compared to training an AHC from scratch on the new dataset.

\subsection{Joint Search Space}
\label{ssec: space}
We focus on the automated design of ST-blocks that extract spatial and temporal features and thus are the core components of CTS forecasting models. 
The joint search space considers two aspects of ST-blocks: 1) the architecture, including operators and their connections, and 2) the hyperparameters, including architecture-related structural hyperparameters (e.g., the hidden dimension) and optimization-related training hyperparameters (e.g., the dropout rate).
Next, we introduce the search spaces of the architecture and hyperparameters in turn, and then show how to combine these into a joint search space.

\subsubsection{Architecture Search Space}
In an ST-block, S/T-operators extract spatial/temporal features, and the connections between operators control the information flow.

\noindent\textbf{Candidate operators.}
By empirically analyzing manually designed CTS forecasting models and the search spaces of existing automated CTS forecasting frameworks, we include two compelling candidate T-operators.
The Gated Dilated Causal Convolution (GDCC)~\cite{DBLP:conf/ijcai/WuPLJZ19,pan2019urban,wu2022autocts} can effectively capture {\it short-term} temporal dependencies. In contrast, Informer (INF-T) \cite{haoyietal-informer-2021}, which is a variant of the Transformer, excels at learning {\it long-term} temporal dependencies. 

We also include two S-operators for extracting two sorts of spatial features. The Diffusion Graph Convolution Network (DGCN) \cite{li2018dcrnn_traffic}, as demonstrated in many popular CTS forecasting studies \cite{DBLP:conf/ijcai/WuPLJZ19,pan2019urban,wu2022autocts}, is effective at capturing static spatial correlations. In addition, the Informer (INF-S) \cite{haoyietal-informer-2021} included due to its strength at discovering dynamic spatial correlations. 

%
We also include an ``identity'' operator to support skip-connections between nodes.
In this way, we obtain a candidate operator set $O$ composed of the above five operators. {The framework can easily accommodate additional operators. Specifically, to add a new operator, we first include the operator in the candidate operator set $O$. Then, we sample some arch-hypers that include the new operator and use them to generate additional clean and noisy samples to retrain the AHC (see Section~\ref{ssec: AHC}). 
The clean and noisy samples collected before can be reused when retraining the AHC, and AHC training is quite efficient (see Section~\ref{ssec: proxy}). 
}

\noindent\textbf{Topological connections.} After selecting the candidate operators, we consider the possible topological connections among the operators within an ST-block.
An ST-block can be represented as a directed acyclic graph (DAG) $G_d$ (e.g., Figure~\ref{fig: gin} left) with $C$ nodes, where each node $h_i$ represents a feature representation and each edge represents an operator $O_i$. 
We propose the following topological connection rules to generate candidate ST-blocks. (1) There is at most one edge from node $h_i$ to node $h_j$, and no edge is allowed from node $h_j$ to node $h_i$, where $i<j$. This is to form the forward flow of a neural network.
(2) The operator on an edge is selected from the chosen candidate set, including the identity operator.


    
    
    

\subsubsection{Hyperparameter Search Space}
We consider two kinds of hyperparameters: structural hyperparameters and training hyperparameters. 
Table~\ref{tab: hy} summarizes the hyperparameters in the hyperparameter search space and also lists their possible values. The framework can easily include additional hyperparameters as well as expanded ranges of values for existing hyperparameters.

\noindent{\bf Structural hyperparameters} relate to the specific structure of an ST-block, including the number $B$ of ST-blocks in a backbone, the number $C$ of nodes in an ST-block, the hidden dimension $H$ of S/T-operators, 
the output dimension $I$, 
%
and the output mode $U$ of an ST-block. 
Larger $B$, $C$, $H$, and $I$ generally result in more expressive ST-blocks but also yield models that are more prone to overfitting on small datasets. 
The output mode $U$ is a binary value indicating which node in an ST-block produces the output. We consider two alternative modes: one takes the last node $h_{C-1}$ as the output, like AutoCTS~\cite{wu2022autocts}, and the other takes the sum of all nodes $h_1$, $h_3$, ..., $h_{C-1}$ as the output, like Graph WaveNet~\cite{DBLP:conf/ijcai/WuPLJZ19}.

\noindent{\bf Training hyperparameters} include the dropout rate $\delta$, which can be used to alleviate overfitting when training a deep CTS model.
The value of $\delta$ can be 0 or 1, which means dropout is used or not used. 
A set of chosen hyperparameter values from the hyperparameter search space can be represented as a $r$-dimensional vector ($r=6$ in this paper). For example, in Table~\ref{tab: hy}, [2, 5, 32, 64, 0, 0] is a possible hyperparameter vector.

\begin{table}[ht]
\small
    \centering
    \caption{The Hyperparameter Search Space.}
    \begin{tabular}{l|l}
    \toprule  
    \emph{Hyperparameters}&\emph{Possible values} \cr
    \hline
    
    \textbf{B (number of ST-blocks)}&{\{2, 4, 6\}} \cr
    \hline

    \textbf{C (number of nodes in an ST-block)}&{\{5, 7\}}
    \cr
    \hline
    
    \textbf{H (hidden dimension)}&{\{32, 48, 64\}}
    \cr
    \hline
    
    \textbf{I (output dimension)}&{\{64, 128, 256\}} 
    \cr
    \hline
    
    \textbf{U (output mode)}&{\{0, 1\}} 
    \cr
    \hline
    
    \textbf{$\delta$ (dropout)}&{\{0, 1\}} 
    \cr
    
    \bottomrule
    \end{tabular}
    \label{tab: hy}
\end{table}


\subsubsection{Encoding of the Joint Search Space}
Having designed the architecture and hyperparameter search spaces, we combine them to construct a joint search space to support the search for an optimal arch-hyper. 
{Performing} a naive combination is infeasible as the two search spaces have different types of encodings (i.e., DAGs vs. vectors). We therefore choose to design the joint search space as a joint dual DAG.
This encoding of the joint search space is easy to explore due to its efficient representation.

\begin{figure}[htb]
  \centering
  \includegraphics[width=0.96\linewidth]{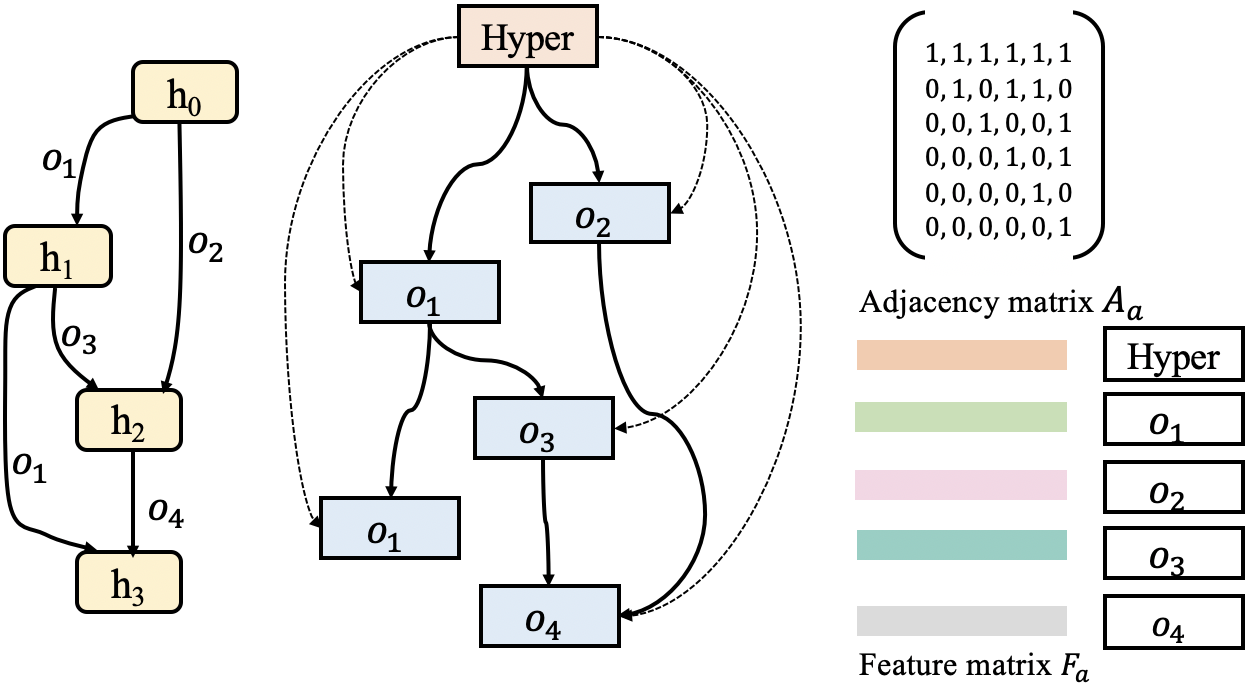}
  \caption{Architecture DAG, Arch-hyper Graph,  and its adjacency matrix and feature matrix representation. $o_i$ represents an operator.
  }
  \label{fig: gin}
\end{figure}

We first convert the original DAG $G_d$ of an architecture (Figure~\ref{fig: gin} left) in the architecture search space into its dual graph $G_d^{\ast}$ (Figure~\ref{fig: gin} middle), where nodes represent operators and edges represent information flow.
This dual form facilitates learning of the representation of an arch-hyper using graph neural networks.
Then, we add a new ``Hyper'' node that represents the hyperparameter setting of the architecture to the dual DAG (Figure~\ref{fig: gin} middle). The ``Hyper'' node connects to all other nodes.
This way, we can use a single DAG $G_a$ (an arch-hyper graph) to represent a complete ST-block containing both the candidate architecture and the hyperparameters, as shown in the middle of Figure~\ref{fig: gin}.

We use an adjacency matrix $A_a$ and a feature matrix $F_a$ to encode an arch-hyper graph $G_a$.
We consider a $G_a$ with $n+1$ nodes ($n=5$ in Figure~\ref{fig: gin} {middle}), where $n$ nodes represent operators and one node represents the hyperparameter settings.
An adjacency matrix $A_a \in\mathbb{R}^{(n+1)\times(n+1)}$ reflects the topology information of $G_a$, where the binary value of an entry ($i$, $j$) indicates whether there is information flow between these two nodes.
We also add self-connections to all nodes.
A feature matrix $F_a\in \mathbb{R}^{(n+1) \times D}$ is also included that contains operator information of each node in an arch-hyper graph.
For the ``Hyper'' node, the original feature is an $r$-dimensional vector from the hyperparameter search space. We first employ min-max normalization to normalize the original feature of the ``Hyper'' node and then convert the normalized feature into a $D$-dimensional embedding:
\begin{align}
F_h=norm(H_o)W_c,
\label{3.1.3.1}
\end{align}
where $H_o \in \mathbb{R}^r$ is the original feature vector of the ``Hyper'' node, $W_c \in \mathbb{R}^{r\times D}$ is a learnable matrix, and $F_h \in \mathbb{R}^D$ is the embedding of the ``Hyper'' node.
For the other $n$ nodes (i.e., the operator nodes), we first embed each operator with an one-hot embedding and then introduce a learnable matrix that converts the one-hot embeddings of all operator nodes into an embedding matrix. Formally, 
\begin{align}
F_e=H_e W_e,
\label{3.1.3.2}
\end{align}
where $H_e\in \mathbb{R}^{n\times |O|}$ and $F_e\in \mathbb{R}^{n\times D}$ are the one-hot embeddings and the transformed embedding matrix of the $n$ nodes, respectively; further, $W_e\in \mathbb{R}^{|O| \times D}$ is the learnable matrix, and $|O|$ is the number of candidate operator types in the architecture search space ($|O|=4$ in Figure~\ref{fig: gin}). 
%
The final feature matrix $F_a\in \mathbb{R}^{(n+1) \times D}$ is the concatenation of the embeddings of the ``Hyper'' node and the operator nodes, i.e., $F_a=concatenate(F_h, F_e)$. 
This way, each arch-hyper in the joint search space can be encoded as an adjacency matrix $A_a$ and a feature matrix $F_a$.
The above learnable parameters $W_c$ and $W_e$ are learned together with the model parameters of the AHC.

\subsection{Architecture-Hyperparameter Comparator}
\label{ssec: AHC}




We propose a novel search framework to find the best arch-hyper in the joint search space. Specifically, we rank arch-hypers according to pairwise comparisons, enabled by an architecture-hyperparameter comparator (AHC), and the top-ranked arch-hypers are selected as the final search result. Unlike in the supernet-based search framework, where a large supernet that embeds all candidate models must reside in memory, in the proposed search framework only a lightweight AHC needs to reside in memory during search. 
This addresses the second limitation of existing automated CTS frameworks, poor scalability.

To rank arch-hypers and to avoid evaluating all candidate pairs, many existing studies build an accuracy estimator, which can be a neural network. The accuracy estimator needs to be trained using a large volume of ($ah$, $R(ah)$) samples, where $ah$ is an arch-hyper and $R(ah)$ is the validation accuracy of a full trained $ah$. 
This way, the accuracy of all candidate arch-hypers in the search space can be estimated, and top-ranked arch-hypers are then selected.
However, it is very time-consuming to collect a large amount of ($ah$, $R(ah)$) samples to train an accuracy estimator, since it is required to fully train many $ah$ to get their validation accuracy $R(ah)$. Further, absolute accuracy is not necessary to achieve the ranking of $ah$; instead, the relative comparison result of two arch-hypers is sufficient to obtain their ranking.

In light of the above considerations, we propose to use a comparator to achieve the relative accuracy relation of two candidate arch-hypers.
Specifically, we design an AHC that takes the embeddings of two arch-hypers ($ah_1$, $ah_2$) as input and outputs a binary value $y$, indicating which arch-hyper may have higher validation accuracy. 
Since the binary relation facilitates obtaining a linear ordering, we can use the AHC to obtain the predicted-accuracy based ranking of arch-hypers in the search space.
Although it is difficult to achieve a fully accurate ranking in practice, we hypothesize that a high-accurate AHC can produce an ordering that approximates the true linear ordering of arch-hypers in the search space, and we offer experimental evidence of this in Section~\ref{ssec: L1L2impact}. 

Given $a$ measured ($ah$, $R(ah)$) pairs, we can build $a(a-1)$ training samples for AHC in the form of ($ah_1$, $ah_2$, $y$) by pairing every two of ($ah$, $R(ah)$) pairs, where $y$ is a binary value indicating which arch-hyper has higher accuracy, thus alleviating the issue of requiring a large amount of training samples. A well trained AHC can then be used to compare all ($ah_1$, $ah_2$) pairs from the joint search space to obtain the magnitude relation between each pair and thus the ranking of candidate arch-hypers.
%

%

\begin{figure}[htb]
  \centering
  \includegraphics[width=0.85\linewidth]{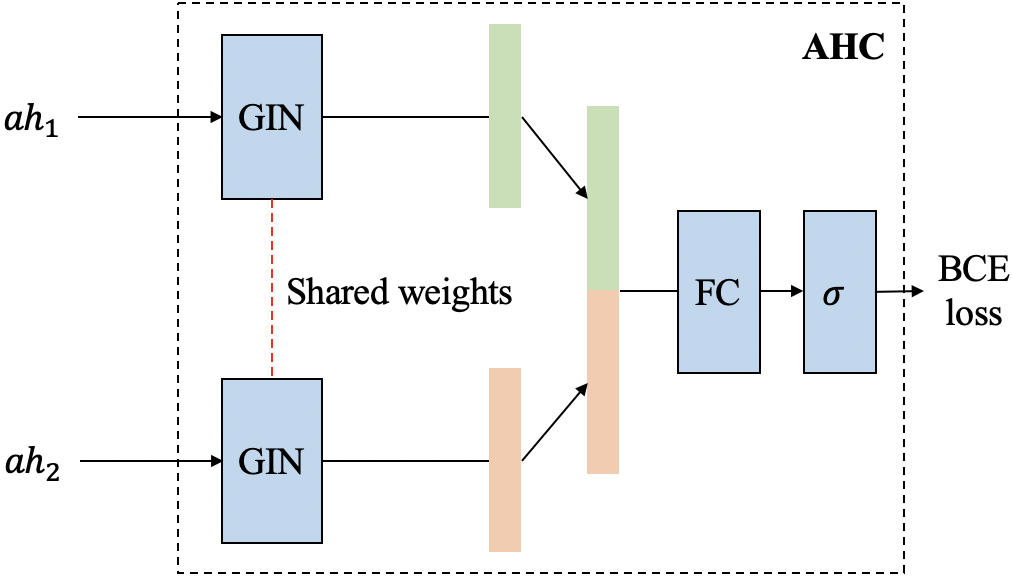}
  \caption{Architecture-Hyperparameter Comparator. 
  }
  \label{fig: AHC}
\end{figure}

Figure~\ref{fig: AHC} shows the proposed AHC. 
The input to the AHC is a pair of arch-hypers ($ah_1$, $ah_2$), each of which is encoded by the adjacency matrix and feature matrix of its joint graph as described in Section~\ref{ssec: space}.
%
Considering the powerful ability of Graph Isomorphism Networks (GINs) to distinguish any two graphs, we leverage GINs to encode the arch-hyper as a compact continuous embedding. Formally, given the adjacency matrix $A_a$ and the feature matrix $F_a$ of an arch-hyper graph, the {corresponding} GIN can be expressed recursively as follows:
\begin{align}
GIN(A_a, F_a) = H^{(L)}
\label{3.2.1}
\end{align}
\begin{align}
H^{(k)} = MLP^{(k)}((1+\epsilon^{(k)})\cdot H^{(k-1)} + AH^{(k-1)}), k=1,2,...,L,
\label{3.2.2}
\end{align}
where $L$ is the number of GIN layers, $H^{(0)}=F_a$, $\epsilon$ is a trainable bias, and $MLP$ is a multi-layer perceptron.
To simplify the representation of an arch-hyper, we use the latent representation of the ``Hyper'' node from $H^{(L)}$ as the representation of the entire arch-hyper, denoted as $l_a$, since the ``Hyper'' node connects to all other nodes in the arch-hyper graph.


%
We encode two input arch-hypers $ah_1$ and $ah_2$ as $l_a$ and $l_a^{\prime}$, respectively, using the same GIN and then concatenate them in the feature dimension:
\begin{align}
L_a = concatenate(l_a, l_a^{\prime}).
\end{align}
Finally, we feed $L_a$ into a classifier that is composed of a fully-connected (FC) layer and a Sigmoid function $\sigma(\cdot)$.
We use 0.5 as the threshold to force the output of the classifier to only be 0 (less than the threshold) or 1 (larger than the threshold). Formally,
\begin{align}
output = \mathbbm{1}(\sigma(FC(L, w_l)) \geq 0.5),
\end{align}
where $w_l$ is the parameter matrix of the fully-connected layer, and $\mathbbm{1}(\cdot)$ is an indicator function with $\mathbbm{1}(\mathrm{X})=1$ if $\mathrm{X}$ is true, and $\mathbbm{1}(\mathrm{X})=0$ otherwise.
We say that $R(ah) \geq R'(ah)$ holds if the output of the AHC is 1, and that $R(ah) < R'(ah)$ holds if the output of the AHC is 0.
We optimize $W_c$, $W_e$, and the parameters of the AHC with Binary Cross-Entropy (BCE) loss.

\subsection{Training AHC with Noisy Proxies}
\label{ssec: proxy}

Training a reliable AHC requires a large number of training samples of the form $(ah_1, ah_2,y)$, where $ah_1$ and $ah_2$ are two arch-hypers, and $y=\mathbbm{1}(R(ah_1)\geq R(ah_2))$ with $R(ah)$ being the forecasting accuracy of an arch-hyper $ah$ on the validation set. However, it is costly to achieve $R(ah)$ since it is required to train an $ah$ from scratch to achieve its forecasting accuracy.
Furthermore, we only need to compare $R(ah_1)$ and $R(ah_2)$ so an exact forecasting accuracy is not essential. 
Therefore, a reasonable and cheaper alternative is a lightweight proxy $R^{\prime}(\cdot)$ that when $R^{\prime}(ah_1) \geq R^{\prime}(ah_2)$ holds, $R(ah_1) \geq R(ah_2)$ holds with high confidence. 
In other words, we aim to find a proxy $R^{\prime}(\cdot)$ that can generate pseudo-labels $y^{\prime}=\mathbbm{1}(R^{\prime}(ah_1)\geq R^{\prime}(ah_2))$ to approximate the true label $y$ for $(ah_1, ah_2)$.
%


This way, we are supposed to design a computation-efficient $R^{\prime}(\cdot)$, so as easily obtain a large number of noisy training samples $(ah_1, ah_2, y^{\prime})$. We call them noisy training samples because $y^{\prime}$ may be incorrectly labelled sometimes. 
%
In this subsection, we first propose an effective proxy $R^{\prime}(\cdot)$
and then describe how to train the AHC in a noise reduction manner to reduce the negative effects introduced by the noisy training samples produced by the proxy.
\subsubsection{Proxy performance metric}
Recent studies have proposed several proxy metrics to quickly measure the performance of neural networks without full training, such as the number of parameters (nparam)~\cite{ning2021evaluating}, Synflow~\cite{abdelfattah2020zero}, Snip~\cite{abdelfattah2020zero}, and neural tangent kernel (NTK) score~\cite{chen2021neural}.
These metrics can be used to generate a pseudo-label $y^{\prime}$ for each $(ah_1, ah_2)$ pair, and consequently generate noisy training samples of the form $(ah_1, ah_2, y^{\prime})$.
However, these proxy metrics are tailored for the computer vision domain, whose operators and model architectures are very different from ours, and thus may not be suitable for CTS forecasting.
Therefore, the noisy training samples $(ah_1, ah_2, y^{\prime})$ generated by these proxies can be severely mislabeled and too noisy to achieve a reliable AHC.

Here, we propose a computation-efficient proxy metric specifically for CTS forecasting, which we later empirically demonstrate to provide a better approximation to $y$ than the aforementioned proxy metrics.
During experiments, we notice that if the validation accuracy of an arch-hyper is higher than another in the first few epochs, then its final validation accuracy is very likely to be higher as well.
Based on this observation, we propose to train an arch-hyper for only $k$ ($k\leq5$) epochs and use the validation accuracy of the under-trained model as a proxy. We call the proposed performance proxy metric as \textit{early-validation proxy} that:
\begin{align}
R^{\prime}(ah)=\mathit{ErrorMetric}(\mathcal{F} (ah)_k,\mathcal{D}_{val}),
\label{eq: evp}
\end{align}
where $\mathcal{F} (ah)_k$ is the CTS forecasting model under arch-hyper setting $ah$ with only $k$ epochs training.

Next, we conduct experiments on two CTS forecasting datasets, PEMS04 and PEMS08, to demonstrate that the \textit{early-validation proxy} is superior to existing performance proxy metrics.
Specifically, for each dataset, we collect $M$ ($M$=50 in this experiment) ($ah$, $R(ah)$) samples by fully training each $ah$, and then pair up them to build $M \times (M-1)$ $(ah_1, ah_2, y)$ samples, where $y=\mathbbm{1}(R(ah_1)\geq R(ah_2))$.
Then we use the proposed and existing proxy metrics to generate pseudo-label $y^{\prime}$ for each $(ah_1, ah_2)$, and count how many samples are wrongly labeled. We propose the \textit{pair-wise ranking accuracy} (PRA) to measure the performance of each proxy that:
\begin{align}
PRA(R^{\prime})=\frac{\sum_{m=1,m\neq n}^{M}\sum_{n=1}^{M}\Delta (R^{\prime}(ah_n),R^{\prime}(ah_m))}{M(M-1)},
\label{eq: pra}
\end{align}
where $\Delta (R^{\prime}(ah_n),R^{\prime}(ah_m))$ is 1 when $R^{\prime}(ah_n) \geq R^{\prime}(ah_m)$ and $R(ah_n) \geq R(ah_m)$, or 0 when $R^{\prime}(ah_n)< R^{\prime}(ah_m)$ and $R(ah_n)< R(ah_m)$, otherwise $\Delta (R^{\prime}(ah_n),R^{\prime}(ah_m))$ is 0.



The pair-wise ranking accuracy for different proxies is shown in Table~\ref{tab: proxy}. It is clear that the proposed proxy metric has significantly higher accuracy than existing proxy metrics on both datasets. Thus, the proposed proxy metric is more suitable for generating noisy training samples for the AHC.

\begin{table}[ht]
\small
    \centering
    \caption{Comparison between the proposed and existing proxy metrics on \textit{pair-wise ranking accuracy}.}
    \begin{tabular}{c|c|c|c|c|c}
        \hline
        &NTK&Synflow&Snip&nParam&Ours(k=5) \cr
        \hline
    PEMS04&{0.64}&{0.57}&{0.49}&0.59&\textbf{0.80} \cr
    PEMS08&{0.69}&0.65&0.50&{0.63}&\textbf{0.82} \cr
    \hline
    \end{tabular}
    \label{tab: proxy}
\end{table}

\subsubsection{Training AHC with denoising algorithms}
While we can use the \textit{early-validation proxy} metric to generate a large number of noisy training samples in a highly efficient manner, it is far from enough to achieve a reliable AHC. 
This is because some noisy samples are wrongly labeled, and the AHC trained with these noisy samples may not be accurate.
To minimize the impact of training with noisy samples, we fully train a few additional arch-hypers to {obtain} clean training samples of the form $(ah_1, ah_2, y)$ and then train the AHC with both noisy and clean samples in a denosing manner.
Formally, given a noisy training set $S_1$ containing $|S_1|$ samples $(ah_1, ah_2,y^{\prime})$ and a clean training set $S_2$ containing $|S_2|$ samples $(ah_1, ah_2,y)$ ($|S_1|\gg |S_2|$), our goal is to train an AHC with $S_1\bigcup S_2$ better than training with only $S_1$ or $S_2$.

There are many alternative methods for training with both noisy and clean labels and have been shown to achieve promising performance on a variety of tasks~\cite{zhang2020distilling,shu2019meta,ren2018learning,DBLP:conf/sigmod/KangAPBZ22}. 
We adopt a commonly used denoising training method, fine-tuning, in our framework. Note that our framework supports the use of other denoising training methods, and we use fine-tuning in this paper for its simplicity and effectiveness.
Specifically, we first randomly sample $L_1$ arch-hypers from the joint search space and use the proposed proxy metric to obtain the proxy score $R^{\prime}(ah)$ for each arch-hyper $ah$. Then we pair up these arch-hypers to produce $L_1(L_1-1)$ noisy samples of the form $(ah_1, ah_2,y^{\prime})$. 
Next, we randomly sample $L_2$ ($L_2<<L_1$) arch-hypers and train them completely to obtain the validation accuracy $R(ah)$ for each arch-hyper $ah$. Then we also pair up these arch-hypers to produce $L_2(L_2-1)$ clean samples of the form $(ah_1, ah_2,y)$.
Note that the processes of collecting noisy and clean samples are independent and can be highly parallelized.
After collecting noisy and clean samples, we first warm up the AHC for $k_t$ epochs using the noisy samples, and then use the clean samples to finetune the AHC until convergence.


\subsection{Search Strategy and AHC Transfer}
\label{ssec: transfer}
\subsubsection{Search strategy}
When a well-trained AHC is achieved, the process for searching the optimal arch-hyper starts. Since the search space is enormous, it is inefficient to compare all candidate hyper-archs using the AHC to get the optimal one. To address this issue, we first shrink the joint search space by removing obviously ineffective arch-hypers based on domain knowledge. Specifically, we remove the arch-hypers that do not contain either spatial or temporal operators since existing works~\cite{DBLP:conf/ijcai/WuPLJZ19,wu2020connecting} demonstrate that considering only temporal or spatial dependencies leads to poor forecasting performance. After that, we consider a heuristic approach, e.g., evolutionary algorithm~\cite{guo2020single}, to find the best arch-hyper in the shrunk joint search space. Specifically, we first sample $K_s$ arch-hypers, which are paired up to produce $K_s(K_s-1)/2$ comparison pairs of the form $(ah_1, ah_2)$, and the descending ranking of the $K_s$ arch-hypers can be easily obtained based on the comparative performance determined by the trained AHC.
{Then, we select the top $k_p$ from the $K_s$ arch-hypers in descending order as the initial population. Each arch-hyper has crossover and mutation probability $p_1$ and $p_2$, respectively, when generating new offspring in each evolution step. The offspring are added to the population, and the learned AHC is used to compare arch-hypers in the population and to remove inferior arch-hypers to keep the population size at $k_p$.
}
Lastly, we choose the top-$K$ arch-hypers from the population to collect their exact forecasting accuracy and pick the one with the highest accuracy as the final searched ST-block. 
The reason we choose top-$K$ arch-hypers instead of the top-1 is that the achieved ranking is not perfectly aligned with the true ranking, thus the top-1 arch-hyper may not be the best arch-hyper. 
In addition, we also note that a random search strategy could replace the evolutionary algorithm for simplicity \cite{bender2018understanding}. 

\subsubsection{Transfer a well-trained AHC}
A practical goal of deep learning for real-world tasks is to \textit{learn features that are transferable}. For example, a pretrained model (e.g., ResNet50) on ImageNet is commonly used as the initialization for finetuning on downstream tasks or different datasets. Naturally, we expect that a well-trained AHC can be an effective initialization for searching on unseen datasets.
Although an arch-hyper performs differently on different CTS forecasting datasets, we observe that the relative performance of two models depends not only on the data, but also on the arch-hyper itself.
For example, Graph WaveNet~\cite{DBLP:conf/ijcai/WuPLJZ19} consistently outperforms STGCN~\cite{yu2018spatio} on different CTS forecasting datasets, demonstrating that the knowledge of the superiority of Graph WaveNet keeps consistent across datasets. 
In light of the above observations, designing an automated framework with transferability is essential and feasible for practical CTS forecasting applications. 
Unlike existing automated CTS forecasting frameworks with poor transferability, our well-trained AHC can be transferred to unseen datasets to further improve the efficiency of the proposed framework.


Specifically, given a well-trained AHC $\mathcal{N}_s$ on a {\it source dataset} $D_s$ (e.g., we used $L_1$ noisy samples and $L_2$ clean samples for training), a new well-trained AHC $\mathcal{N}_t$ on a {\it target dataset} $D_t$ can be obtained by {\it transferring} $\mathcal{N}_s$ from $D_s$ to $D_t$ with much less arch-hyper samples $ah_t$ on $D_t$:
\begin{equation}
    \mathcal{N}_t(D_t) \xleftarrow[]{ah_t}  \mathcal{N}_s(D_s),
\end{equation}
where $ah_t$ consists of $z_1$ noisy samples and $z_2$ clean samples, and $z_1 \ll L_1, z_2 \ll L_2$. 
The transfer process is particularly implemented using the fine-tuning technique. Our AHC transfer strategy can overcome the {\it one-time use} shortcoming of existing automated frameworks and significantly improve efficiency. 
We experimentally demonstrate that the proposed transfer method can significantly improve search efficiency without loss of search accuracy compared to training our framework on new datasets from scratch. The complete search strategy, including the optional transfer, is shown in Algorithm~\ref{alg}.

\begin{algorithm}[htb]
\caption{Search Algorithm}
\begin{flushleft}
{\bf Input:} 
target CTS dataset $\mathcal{D}$, joint search space $\Omega$, AHC $\mathcal{N}_s$ on a source dataset $\mathcal{D}_s$ (optional), $\{L_1, L_2, z_1, z_2\}$ s.t. $L_1\gg L_2$, $L_1\gg z_1$, $L_2\gg z_2$;

{\bf Output:}
optimal arch-hyper $ah^*$
\end{flushleft}
\begin{algorithmic}[1]

\State Split $\mathcal{D}$ into $\mathcal{D}_{train}$, $\mathcal{D}_{val}$, and $\mathcal{D}_{test}$
\State {\bf if} $\mathcal{N}_s$ exists:
\State \hspace{0.1in} Randomly sample $s_{z_1}=\{ah\in\Omega\}$, $|s_{z_1}|=z_1$
\State \hspace{0.1in} Generate $S_{z_1}=\{(ah_1, ah_2, y^{\prime})|ah_*\in s_{z_1}\}$, $|S_{z_1}|=z_1(z_1-1)$
\State \hspace{0.1in} Randomly sample $s_{z_2}=\{ah\in\Omega\}$, $|s_{z_2}|=z_2$
\State \hspace{0.1in} Generate $S_{z_2}=\{(ah_1, ah_2, y)|ah_*\in s_{z_2}\}$, $|S_{z_2}|=z_2(z_2-1)$
\State \hspace{0.1in} {\bf while} not converged {\bf do}
\State \hspace{0.2in}  $\mathcal{N}_t$=fine-tuning($\mathcal{N}_s, {S_{z_1}}, {S_{z_2}}$)
\State \hspace{0.1in} {\bf end}
\State {\bf else }
\State \hspace{0.1in} Randomly sample $s_1=\{ah\in\Omega\}$, $|s_1|=L_1$
\State \hspace{0.1in} Generate $S_1=\{(ah_1, ah_2, y')|ah_*\in s_1\}$, $|S_1|=L_1(L_1-1)$
\State \hspace{0.1in} Randomly sample $s_2=\{ah\in\Omega\}$, $|s_2|=L_2$
\State \hspace{0.1in} Generate $S_2=\{(ah_1, ah_2, y)|ah_*\in s_2\}$, $|S_2|=L_2(L_2-1)$
\State \hspace{0.1in} Initialize $\mathcal{N}_t$
\State \hspace{0.1in} {\bf for} t = 1, ..., $k_t$ {\bf do}
\State \hspace{0.2in} Train $\mathcal{N}_t$ with $S_1$.
\State \hspace{0.1in} {\bf end for}
\State \hspace{0.1in} {\bf while} not converged {\bf do}
\State \hspace{0.2in} Train $\mathcal{N}_t$ with $S_2$
\State \hspace{0.1in} {\bf end}
\State {\bf end if}
%
\State Shrink $\Omega$ to $\Omega_s$ with domain knowledge
\State Heuristic search and rank $ah$ $\in$ $\Omega_s$ using the trained AHC.
\State Train and evaluate the top-$K$ $ah\in\Omega_s$ on $\mathcal{D}_{train}$ and $\mathcal{D}_{val}$, respectively
\State {\bf return} The optimal $ah^*$ that yields the highest validation accuracy
\end{algorithmic}
\label{alg}
\end{algorithm}

\section{Experiments}
\label{sec: exp}

We conduct experiments on four CTS datasets for multi-step forecasting and two CTS datasets for single-step forecasting. The results demonstrate that our proposed framework successfully addresses the three limitations of previous work with higher forecasting accuracy, lower memory consumption, and faster search process. 


\subsection{Experimental Settings}
\label{ssec:expsetting}

\subsubsection{Datasets}
To enable easy and fair comparisons, we use widely adopted benchmark datasets for single- and multi-step forecasting experiments.
{We consider both kinds of forecasting for two reasons. First, the literature on CTS forecasting often considers both of these---thus, considering both enables fair comparisons.
Second, the two kinds of forecasting focus on evaluating different aspects of the abilities of CTS models, with multi-step forecasting typically focusing on evaluating the ``average'' ability to capture long-term and short-term temporal dependencies, while single-step forecasting focuses on evaluating the ability to capture either long-term or short-term dependencies, depending on the time length of a single step.}

\noindent
\textbf{Multi-step forecasting: }
\begin{itemize}[leftmargin=*]
    \item PEMS03, PEMS04, PEMS07 and PEMS08~\cite{song2020spatial}: These four datasets record the traffic flow in four different regions of California, and are collected from the Caltrans Performance Measurement System (PeMS).
    Each dataset contains traffic flow data of hundreds of roads, and the traffic flow readings are aggregated into 5-minute windows, resulting in 12 data points per hour.
    We construct an adjacency matrix reflecting the correlation among roads for each dataset based on pairwise road network distances~\cite{song2020spatial}. 
\end{itemize}
We summarize the statistics of the four datasets in Table~\ref{tab: data}, where $N$ represents the number of time series, $T$ represents the total number of timestamps, and ``Split Ratio'' refers to the ratio between train, validation, and test sets.
We consider two different settings of input ($P$) and forecasting ($Q$) lengths, i.e., $P$-12/$Q$-12 and $P$-48/$Q$-48, corresponding to short- and long-term CTS forecasting, respectively.

%

\noindent
\textbf{Single-step forecasting: }
\begin{itemize}[leftmargin=*]
    \item Solar-Energy~\cite{lai2018modeling}: This dataset contains the solar power production records collected from 137 PV plants in the Alabama State.
    \item Electricity~\cite{lai2018modeling}: This dataset contains the electricity consumption records collected from 321 clients.
\end{itemize}

We also summarize the statistics of the two datasets in Table~\ref{tab: data}. Following existing literature~\cite{lai2018modeling,shih2019temporal,wu2020connecting}, we use the past 168 timestamps to predict the future 1 timestamp, where the future 1 timestamp can be the 3rd or 24th timestamp.
{Note that the single-step datasets do not include spatial information that enables computing the distances among sensors. Thus, there is no predefined distance-based adjacency matrix, which also means that the strengths of spatial correlations between time series in the single-step datasets are unknown in advance.}

\begin{table}[ht]
\small
    \centering
    \caption{Dataset statistics. Split ratio refers to the train-validation-test split ratio. P/Q refers to the input length and the forecasting length.}
    \begin{tabular}{l l l l l l}
        \hline
         Dataset & $N$ & $T$ & Split Ratio & $P$/$Q$ & $P$/$Q$\\
        \hline\hline
         PEMS03 & 358 & 26,208 & 6:2:2 & 12/12 & 48/48 \\
         PEMS04 & 307 & 16,992 & 6:2:2 & 12/12 & 48/48 \\
         PEMS07 & 883 & 28,224 & 6:2:2 & 12/12 & 48/48 \\
         PEMS08 & 170 & 17,856 & 6:2:2 & 12/12 & 48/48 \\
         Solar-energy & 137 & 52,560 & 6:2:2 & 168/1 (3rd) & 168/1 (24th) \\
         Electricity & 321 & 26,304 & 6:2:2 & 168/1 (3rd) & 168/1 (24th) \\
        \hline
    \end{tabular}
    \label{tab: data}
\end{table}

\subsubsection{Evaluation Metrics}
To compare the performance of different CTS forecasting models, we follow previous studies~\cite{yu2018spatio,li2018dcrnn_traffic,DBLP:conf/ijcai/WuPLJZ19,bai2020adaptive,wu2020connecting} to use mean absolute error (MAE), root mean squared error (RMSE), and mean absolute percentage error (MAPE) for estimating the accuracy of multi-step forecasting, and use Root Relative Squared Error (RRSE) and Empirical Correlation Coefficient (CORR) for estimating the accuracy of single-step forecasting. 
For MAE, RMSE, MAPE, and RRSE, smaller values are better, while larger values for CORR are better.
To evaluate the performance of the AHC, we use Spearman's rank correlation coefficient ($\rho$) to measure the similarity of the rankings produced by the AHC to the true ranking of arch-hypers.

\subsubsection{Baselines}
In recent works, we have seen that automated methods often outperform manually designed methods, thus we compare the proposed framework with two best-performing manually designed CTS forecasting models and two state-of-the-art automated methods.
%
The results of the baselines are obtained by running the originally released source code.

\begin{itemize}[leftmargin=*]
    \item[$\bullet$] MTGNN: A multivariate time series forecasting model, which employs mix-hop graph convolution and dilated inception convolution to build ST-blocks~\cite{wu2020connecting}.
    \item[$\bullet$] AGCRN: Adaptive graph convolutional recurrent network, which employs 1D GCN and GRU to build ST-blocks~\cite{bai2020adaptive}.
    \item[$\bullet$] AutoSTG: A supernet-based automated CTS forecasting framework, which employs DGCN and 1D convolution to build the search space, and introduces meta learning to learn the weights of neural operators~\cite{pan2019urban}. However, it relies on the coordinates of sensors to build the attributed graph to work, which are not available on the evaluation datasets, so we replace them with the ordinal number of time series during experiment.
    \item[$\bullet$] AutoCTS: A supernet-based automated CTS forecasting framework, which focuses on selecting the optimal set of neural operators to build the search space~\cite{wu2022autocts}.
\end{itemize}

\subsubsection{Implementation Details}
We have implementation for both preparing AHC and training the CTS forecasting model.

\noindent \textbf{Setting up the AHC.}
Since we allow ST-blocks to have different numbers of nodes, the shape of the adjacency matrix $A_a$ may be different for different arch-hypers. Thus, we pad the size of the adjacency matrix to 14 with zero paddings.
We set the number $L$ of layers of the GIN to 4, with $D=128$ hidden units in each layer.
For training the AHC, we use Adam~\cite{kingma2014adam} with a learning rate of 0.0001 and a weight decay of 0.0005 as the optimizer. The batch size is set to 8.
For pretraining the AHC on the source dataset, we first use $L_1=2000$ arch-hypers to generate noisy samples to train the AHC for $k_t=10$ epochs, with an early stopping patience of 3 epochs; then we use $L_2=150$ arch-hypers to generate clean samples to finetune the AHC for 10 epochs, with an early stopping patience of 3 epochs.
For transferring the AHC to a target dataset, we only use $z_1=100$ arch-hypers to generate noisy samples and $z_2=5$ arch-hypers to generate clean samples to finetune a trained AHC for 3 epochs.

%
{We set the crossover and mutation probability $p_1$ and $p_2$ to 0.8 and 0.2, respectively, and the population size $k_p$ to 10. Lastly, we choose the top-3 arch-hypers from the population.}


%

\noindent \textbf{Setting up CTS forecasting models.}
We use MAE as the training objective to train CTS forecasting models, and use Adam with a learning rate of 0.001 and a weight decay of 0.0001 as the optimizer. The batch size is set to 64.
%
%
For generating clean samples, we set the training epochs to 100.
For generating noisy samples, we set the training epochs $k$ to 5.

%

\noindent \textbf{Reproducibility. }
To support the reproducibility, we provide the source code and links to the datasets in the supplementary materials. Source code will be released upon acceptance. We conduct all experiments on multiple Nvidia Quadro RTX 8000 GPUs.

\subsection{Experimental Results}
\subsubsection{Performance Comparison}
Table~\ref{tab: main} presents the performance comparison between our framework and the baselines on the four multi-step CTS forecasting datasets.
To demonstrate the robustness of our framework to different source datasets for AHC transfer, we report the results of two specific settings {\it Ours-1} and {\it Ours-2}, which use PEMS08 and PEMS04 as the source datasets, respectively.
Since the baselines do not manually tune hyperparameters under the $P$-48/$Q$-48 setting, for fair comparison, we conduct grid-search for them to find the best hidden dimension $H$ and the output dimension $I$, and also include the hyperparameter setting they use under the $P$-12/$Q$-12 setting.
To facilitate observation, we use {\bf bold} and \underline{underline} to highlight the best and the second best results, respectively. We summarize main observations as follows.

\begin{table*}[ht]
\small
    \centering
    \caption{Performance Comparison of Multi-step Forecasting.} %
    
    \begin{tabular}{c|c|cccccc||cccccc}
        \hline
        \multirow{2}{*}{Data}&
        \multirow{2}{*}{Metric}&
        \multicolumn{6}{c||}{$P$-12/$Q$-12}&
        \multicolumn{6}{c}{$P$-48/$Q$-48} \cr  
        &&MTGNN&AGCRN&AutoSTG&{AutoCTS}&\emph{Ours-1}&\emph{Ours-2}&MTGNN&AGCRN&AutoSTG&{AutoCTS}&\emph{Ours-1}&\emph{Ours-2} \cr
        \hline
    \multirow{3}{*}{PEMS03}&MAE&{15.10}&15.89&17.97&{14.71}&\underline{14.60}&\textbf{14.59}&{20.66}&19.84&22.46&{19.61}&\textbf{18.37}&\underline{18.44} \cr
     &RMSE&{25.93}&28.12&28.47&{24.54}&\underline{24.35}&\textbf{24.21}&{34.13}&33.93&37.57&{32.93}&\underline{30.86}&\textbf{30.77} \cr
     &MAPE&15.67\%&{15.38}\%&{18.08}\%&{14.39}\%&\textbf{13.85}\%&\underline{14.02}\%&{24.31\%}&18.56\%&21.32\%&{19.80\%}&\textbf{17.69}\%&\underline{17.76}\% \cr
    \hline
    \multirow{3}{*}{PEMS04}&MAE&{19.32}&19.83&20.46&{19.13}&\underline{18.97}&\textbf{18.95}&{24.56}&22.83&26.17&{23.69}&\textbf{22.68}&\underline{22.74} \cr
     &RMSE&{31.57}&32.26&32.18&{30.44}&\underline{30.36}&\textbf{30.31}&{37.15}&36.20&41.07&{36.45}&\textbf{35.28}&\underline{35.35} \cr
     &MAPE&13.52\%&{12.97}\%&13.77\%&{12.89}\%&\underline{12.81}\%&\textbf{12.75}\%&{19.35\%}&\textbf{15.09}\%&18.02\%&{18.04\%}&{15.93}\%&\underline{15.82}\% \cr
    \hline
    \multirow{3}{*}{PEMS07}&MAE&22.07&{21.31}&26.77&{20.93}&\textbf{20.65}&\underline{20.77}&{25.74}&24.90&38.56&{25.49}&\underline{23.74}&\textbf{23.53} \cr
     &RMSE&{35.80}&35.06&41.63&{33.69}&\underline{33.54}&\textbf{33.49}&{40.59}&41.48&62.63&{40.33}&\underline{38.69}&\textbf{38.46} \cr
     &MAPE&9.21\%&{9.13}\%&11.63\%&{8.90}\%&\underline{8.81}\%&\textbf{8.76}\%&{11.78\%}&\underline{10.68}\%&18.14\%&{11.63\%}&{10.69}\%&\textbf{10.57}\% \cr
    \hline
    \multirow{3}{*}{PEMS08}&MAE&{15.71}&15.95&16.23&{14.82}&\textbf{14.68}&\underline{14.72}&{20.37}&19.44&21.25&{18.85}&\textbf{17.68}&\underline{17.73} \cr
     &RMSE&{24.62}&25.22&25.72&{23.64}&\underline{23.46}&\textbf{23.43}&{30.75}&31.40&34.59&{29.13}&\textbf{27.95}&\underline{27.98} \cr
     &MAPE&10.03\%&{10.09}\%&{10.25}\%&{9.51}\%&\textbf{9.41}\%&\underline{9.45}\%&{16.69\%}&13.38\%&14.39\%&{15.08}\%&\underline{12.60}\%&\textbf{12.47}\% \cr
    \hline
    \end{tabular}
    \label{tab: main}
\end{table*}

First, our framework, either Ours-1 or Ours-2, achieves consistent state-of-the-art forecasting accuracy on all datasets and $P$/$Q$ settings. 
{In particular, our framework outperforms two existing automated methods, AutoSTG, which employs different S/T operators and a different search algorithm, and AutoCTS, which employs the same S/T operators but a different search algorithm.
}


%
Second, our framework achieves more significant improvements on the $P$-48/$Q$-48 setting than on the $P$-12/$Q$-12 setting. 
This is because the architectures and hyperparameters of the baselines are specifically designed for the $P$-12/$Q$-12 setting in the original studies.
And under the $P$-48/$Q$-48 setting, although we conduct grid-search to find the best hyperparameters for the baselines, it does not yield competitive arch-hypers.
In contrast, our proposed joint search framework can find high-performance arch-hypers under different $P$/$Q$ settings.
This result indicates the necessity of joint search architecture and hyperparameters and the advances of our proposed framework in this context.

%
Third, the results of Ours-1 and Ours-2 are overall comparable. This demonstrates that our AHC transfer method is effective regardless which dataset is used as the source dataset. 
Besides, the results using the transferred AHC (ours-1 on PEMS04 and ours-2 on PEMS08) and the originally trained AHC (ours-2 on PEMS04 and ours-1 on PEMS08) are closely analogous, demonstrating that our AHC transfer method can improve the search efficiency without sacrificing accuracy.

Table~\ref{tab: single} presents the performance comparison between our framework and the baselines on the two single-step CTS forecasting datasets{, where the source datasets of {\it Ours-1} and {\it Ours-2} remain PEMS08 and PEMS04, respectively.}
{
Since we use the same experimental settings, the results of MTGNN and AutoCTS in Table 5 are obtained from the original papers.
Next, as AGCRN and AutoSTG do not report results on the single-step datasets, for a fair comparison, we conduct a grid search to determine the best settings for two key hyperparameters: the hidden dimension $H$ and the output dimension $I$. We keep the settings of other hyperparameters as in the original papers.
Note that AutoSTG relies on a predefined adjacency matrix for spatial graph convolution operators. As such a matrix is not available in the single-step CTS datasets, we drop the spatial graph convolution operators in AutoSTG's search space when running it on the single-step CTS datasets. 
Also note that none of the remaining baselines and our framework require predefined adjacency matrices, as they encompass self-adaptive adjacency matrices to learn correlations among time series.
}
%
We observe similar trends for single-step forecasting and the multi-step forecasting. Specifically, our framework achieves the consistently best performance across all datasets and forecasting timestamps, suggesting the success of our joint search; our searched models with different AHC source datasets, ours-1 and ours-2, have comparable results, indicating that our AHC transfer method is robust to different source datasets.

\begin{table}[htbp]
\small
    \centering
    \caption{Performance Comparison of Single-step Forecasting.}
    \begin{tabular}{c|c|c|c|c|c}
        \toprule  
        \multicolumn{2}{c|}{Data}&
        \multicolumn{2}{c|}{Solar-Energy}&        \multicolumn{2}{c}{Electricity} \cr
    \hline
    Models&Metric&3&24&3&24 \cr
    \midrule 
    \multirow{2}{*}{MTGNN}&RRSE&{0.1778}&{0.4270}&{0.0745}&{0.0953} \cr
     &CORR&{0.9852}&0.9031&{0.9474}&{0.9234} \cr
    \hline
    \multirow{2}{*}{{AGCRN}}&{RRSE}&{0.1830}&{0.4602}&{0.1033}&{0.0994}\cr    &{CORR}&{0.9846}&{0.9016}&{0.8854}&{0.9073} \cr
    \hline
    \multirow{2}{*}{{AutoSTG}}&{RRSE}&{0.2094}&{0.5066}&{0.1188}&{0.0998} \cr &{CORR}&{0.9811}&{0.8611}&{0.9070}&{0.8846}\cr
    \hline
    \multirow{2}{*}{{AutoCTS}}&RRSE&{0.1750}&{0.4143}&{0.0743}&{0.0947} \cr
     &CORR&{0.9855}&{0.9085}&{0.9477}&{0.9239} \cr
    \hline
    \multirow{2}{*}{\emph{Ours-1}}&RRSE&\textbf{0.1657}&\textbf{0.3980}&\underline{0.0736}&\textbf{0.0921} \cr
     &CORR&\textbf{0.9875}&\textbf{0.9152}&\underline{0.9483}&\textbf{0.9253} \cr
    \hline
    \multirow{2}{*}{\emph{Ours-2}}&RRSE&\underline{0.1663}&\underline{0.4009}&\textbf{0.0732}&\underline{0.0935} \cr
     &CORR&\underline{0.9886}&\underline{0.9138}&\textbf{0.9487}&\underline{0.9244} \cr
    \bottomrule
    \end{tabular}
    \label{tab: single}
\end{table}

\subsubsection{Scalability Study}
\begin{figure}[htb]
\center
\subfigure[Memory w.r.t. N]{
\begin{minipage}[c]{0.44\linewidth} 
\centering
\includegraphics[width=\linewidth]{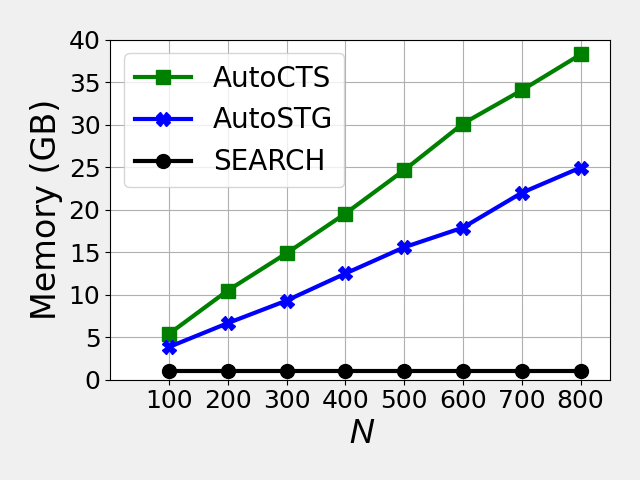}
\label{fig: N}
\end{minipage}
}
\subfigure[Memory w.r.t. P]{
\begin{minipage}[c]{0.44\linewidth} 
\centering
\includegraphics[width=\linewidth]{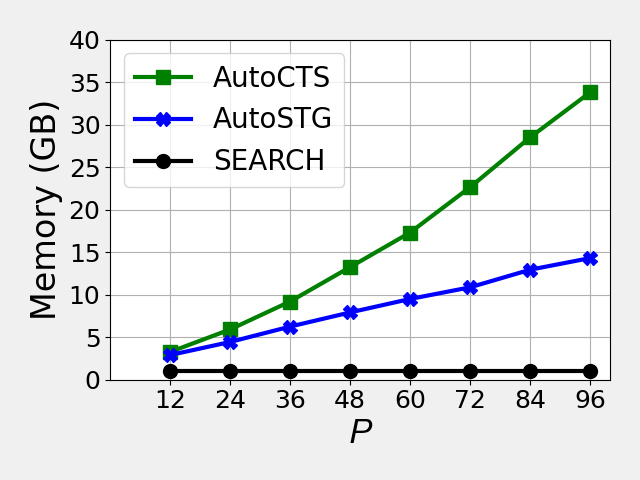}
\label{fig: T}
\end{minipage}
}
\caption{Scalability Comparison on the PEMS07 dataset.}
\label{fig: memory}
\end{figure}

We compare the scalability of our framework and existing automated frameworks and plot the results in Figure~\ref{fig: memory}. The PEMS07 dataset is used for experiments because it has the largest number of time series $N$, which is most suitable for observing how the memory cost increases with $N$.
Specifically, we first fix the number $P$ of input timestamps to 12, and vary $N\in \{100, 200, 300, 400, 500, 600, 700, 800\}$ to observe the scalability w.r.t. $N$;
then, we fix $N$ to 50, and vary $P\in\{12, 24, 36, 48, 60, 72, 84, 96\}$ to observe the scalability w.r.t. $P$.

In Figure~\ref{fig: N}, the memory usage of AutoSTG and AutoCTS grows rapidly with $N$, while the memory usage of our framework is a constant and does not change with $N$. Similarly, as shown in Figure~\ref{fig: T}, the memory usage of AutoSTG and AutoCTS grows rapidly with $P$, while the memory usage of our framework is a small constant and does not change with $P$. This demonstrates that the proposed framework is more scalable than existing automated frameworks.
The reason is that the AHC is implemented with a lightweight graph neural network and is independent of CTS datasets. While the memory cost of existing automated frameworks is positively correlated with $N$ and $P$ of CTS datasets.
The high scalability of our framework allows to work on various sizes of CTS datasets (i.e., a large range of $N$ and $P$), while AutoSTG and AutoCTS may fail because their memory cost goes overflow more easily.


\subsubsection{Transferability Study}

We study the effectiveness of the proposed AHC transfer approach and summarize the results in Table~\ref{tab: transfer1}. The PEMS04 and PEMS08 datasets are used as each other's source and target datasets. We build three variants: (1) {\bf w/ transfer} that uses the proposed AHC transfer approach with $z_1=100$ $(ah, R^{\prime}(ah))$ pairs and $z_2=5$ $(ah, R(ah))$ pairs for fine-tuning; (2) {\bf w/o transfer} that trains an AHC directly on the target dataset using $z_1=100$ $(ah, R^{\prime}(ah))$ pairs and $z_2=5$ $(ah, R(ah))$ pairs without transfer; (3) {\bf random} that randomly selects 3 arch-hypers from the joint search space and trains them completely to get the best one.
{
To eliminate the bias caused by the sampling procedure, we first determine five random seeds.
For {\it w/ transfer} and {\it w/o transfer}, we select arch-hypers to generate noisy and clean samples under each random seed setting (with the same settings for $z_1$ and $z_2$). For {\it random}, we select 3 arch-hypers under each random seed setting. 
For each variant, we collect the results obtained with the five different random seeds and report their mean and standard deviation.
}

%
%
We can observe that for both datasets the {\it w/o transfer} is significantly worse than the {\it w/ transfer} with almost the same performance as the {\it random} variant.
It demonstrates that without AHC transfer, a small number of $(ah, R^{\prime}(ah))$ and $(ah, R(ah))$ pairs is far from enough to achieve a reliable AHC and the search based on such an AHC is just like random. Conversely, given a AHC trained on a source dataset, we can easily obtain a reliable AHC on a target dataset by fine-tuning it with a small number of $(ah, R^{\prime}(ah))$ and $(ah, R(ah))$ pairs, thus ensuring high training efficiency.


\begin{table}[ht]
\small
    \centering
    \caption{Transferability Study on the $P$-12/$Q$-12 Multi-step Forecasting (mean$\pm$standard deviation). 
    } %
    
    \begin{tabular}{c|c|ccc}
        \hline
        {Target}&
        \multirow{2}{*}{Metric}&\multirow{2}{*}{w/ transfer}&\multirow{2}{*}{w/o transfer}&\multirow{2}{*}{random} \cr
        {Dataset}&&&& \cr
        \hline
    \multirow{3}{*}{PEMS04}&MAE&\textbf{18.93$\pm$0.08}&{19.44$\pm$0.16}&{19.36$\pm$0.19} \cr
     &RMSE&\textbf{30.29$\pm$0.11}&{30.62$\pm$0.25}&{30.64$\pm$0.21} \cr
     &MAPE&\textbf{12.76\%$\pm$0.13\%}&{14.08\%$\pm$0.28\%}&{14.39\%$\pm$0.33\%} \cr
    \hline
    \multirow{3}{*}{PEMS08}&MAE&\textbf{14.64$\pm$0.09}&{15.38$\pm$0.24}&{15.32$\pm$0.22} \cr
    &RMSE&\textbf{23.41$\pm$0.15}&{24.19$\pm$0.19}&{24.14$\pm$0.26} \cr
    &MAPE&\textbf{9.44\%$\pm$0.11\%}&{10.26\%$\pm$0.31\%}&{10.68\%$\pm$0.36\%} \cr
    \hline
    \end{tabular}
    \label{tab: transfer1}
\end{table}


\subsubsection{Ablation Studies}
We conduct ablation studies to investigate the effectiveness of each key component of the proposed framework.
{We report results for all datasets and settings in Tables~\ref{tab: ablation_12} to ~\ref{tab: ablation_single}, where the source dataset of {\it Ours-1} remains PEMS08, and similar trends can be observed when using {\it Ours-2}, i.e., using PEMS04 as the source dataset.
}
We compare our framework with the following variants.

\begin{itemize}[leftmargin=*]
    \item[$\bullet$] \textbf{w/o joint} only searches for architectures. 
    We use a fixed hyperparameter setting (4, 5, 32, 256, 1, 0), which is commonly used in previous studies~\cite{DBLP:conf/ijcai/WuPLJZ19,wu2022autocts}, and perform our framework to search for the best architecture.
    \item[$\bullet$] \textbf{w/o clean} does not train an AHC with clean samples but only with noisy samples; 
    \item[$\bullet$] \textbf{w/o noisy} does not train an AHC with noisy samples but only with clean samples;
    \item[$\bullet$] \textbf{w/o denoising} does not train an AHC with the denoising method. Instead, it blends clean and noisy samples for AHC training.
\end{itemize}


\begin{table}[!htbp]
\small
    \centering
    {
    \caption{ Ablation studies, $P$-12/$Q$-12}
    \begin{tabular}{c|c|ccccc}
        \hline
        \multirow{2}{*}{Data}&
        \multirow{2}{*}{Metric}&
        \multirow{2}{*}{\emph{Ours-1}}&
        w/o&w/o&w/o&w/o \cr
        &&&joint&clean&noisy&denosing \cr
        \hline
    \multirow{3}{*}{PEMS03}&MAE&\textbf{14.60}&14.81&14.62&14.74&14.67 \cr
     &RMSE&\textbf{24.35}&24.69&24.39&24.52&24.22 \cr
     &MAPE&\textbf{13.85}\%&14.43\%&14.05\%&14.38\%&14.12\% \cr
    \hline
    \multirow{3}{*}{PEMS04}&MAE&\textbf{18.97}&19.32&19.11&19.24&19.01 \cr
     &RMSE&\textbf{30.36}&30.59&30.38&30.50&30.34 \cr
     &MAPE&\textbf{12.81}\%&{13.05}\%&12.85\%&13.12\%&12.91\% \cr
    \hline
    \multirow{3}{*}{PEMS07}&MAE&\textbf{20.65}&20.98&20.89&20.95&20.83 \cr
     &RMSE&\textbf{33.54}&33.84&33.68&33.72&33.60 \cr
     &MAPE&\textbf{8.81}\%&{8.97}\%&8.92\%&9.06\%&8.85\% \cr
    \hline
    \multirow{3}{*}{PEMS08}&MAE&\textbf{14.68}&15.07&14.83&14.89&14.77 \cr     &RMSE&\textbf{23.46}&24.01&23.58&23.76&23.53 \cr &MAPE&\textbf{9.41}\%&9.79\%&9.64\%&9.81\%&9.52\% \cr
    \hline
    \end{tabular}
    \label{tab: ablation_12}
    }
\end{table}

\begin{table}[!htbp]
\small
    \centering
    {
    \caption{ Ablation studies, $P$-48/$Q$-48.}
    \begin{tabular}{c|c|ccccc}
        \hline
        \multirow{2}{*}{Data}&
        \multirow{2}{*}{Metric}&
        \multirow{2}{*}{\emph{Ours-1}}&
        w/o&w/o&w/o&w/o \cr
        &&&joint&clean&noisy&denosing \cr
        \hline
    \multirow{3}{*}{PEMS03}&MAE&\textbf{18.37}&19.57&18.65&18.97&18.82 \cr
     &RMSE&\textbf{30.86}&32.31&31.15&31.60&31.02 \cr
     &MAPE&\textbf{17.69}\%&{20.13}\%&18.19\%&19.78\%&18.00\% \cr
    \hline
    \multirow{3}{*}{PEMS04}&MAE&\textbf{22.68}&23.75&22.86&23.44&22.92 \cr
     &RMSE&\textbf{35.28}&36.67&35.41&35.84&35.37 \cr
     &MAPE&\textbf{15.93}\%&{17.64}\%&16.18\%&16.93\%&16.02\% \cr
    \hline
    \multirow{3}{*}{PEMS07}&MAE&\textbf{23.74}&25.19&23.94&24.46&23.89 \cr
     &RMSE&\textbf{38.69}&39.95&38.73&39.31&38.57 \cr
     &MAPE&\textbf{10.69}\%&{11.87}\%&10.75\%&11.04\%&10.62\% \cr
    \hline
    \multirow{3}{*}
    {PEMS08}&MAE&\textbf{17.68}&{18.94}&18.02&18.31&17.82 \cr
    &RMSE&\textbf{27.95}&29.18&28.22&28.54&28.15 \cr
    &MAPE&\textbf{12.60}\%&{14.29}\%&12.85\%&13.51\%&12.73\% \cr
    \hline
    \end{tabular}
    \label{tab: ablation_48}
    }
\end{table}

\begin{table}[!htbp]
\small
    \centering
    {
    \caption{ Ablation studies, single-step datasets.}
    \begin{tabular}{c|c|ccccc}
        \hline
        \multirow{2}{*}{Data}&
        \multirow{2}{*}{Metric}&
        \multirow{2}{*}{\emph{Ours-1}}&
        w/o&w/o&w/o&w/o \cr
        &&&joint&clean&noisy&denosing \cr
        \hline
    {Solar-Energy}&RRSE&\textbf{0.1657}&0.1762&0.1669&0.1691&0.1677 \cr
     (3)&CORR&\textbf{0.9875}&0.9848&0.9881&0.9869&0.9874 \cr
    \hline
    {Solar-Energy}&RRSE&\textbf{0.3980}&0.4138&0.4033&0.4051&0.4010 \cr
     (24)&CORR&\textbf{0.9152}&0.9087&0.9118&0.9106&0.9132 \cr
    \hline
    {Electricity}&RRSE&\textbf{0.0736}&0.0752&0.0741&0.0748&0.0739 \cr
     (3)&CORR&\textbf{0.9483}&0.9441&0.9478&0.9462&0.9480 \cr
    \hline
    {Electricity}&RRSE&\textbf{0.0921}&0.0956&0.0942&0.0954&0.0938 \cr
     (24)&CORR&\textbf{0.9253}&09223&0.9236&0.9228&0.9242 \cr
    \hline
    \end{tabular}
    \label{tab: ablation_single}
    }
\end{table}


%
From Tables~\ref{tab: ablation_12} to ~\ref{tab: ablation_single},
we can observe that (1) our proposed framework consistently outperforms all variants on all evaluation metrics that every tested component is essential to the success of the proposed framework;
(2) {\it w/o joint} is much worse than {\it Ours-1} that it is necessary to jointly search for architectures and hyperparameters to ensure satisfactory performance; 
%
(3) both {\it w/o clean} and {\it w/o noisy} are worse than {\it Ours-1} that both noisy samples and clean samples are keys to a reliable AHC thus a competitive searched model; Clean samples are essential to alleviating the bias produced by noisy samples, while noisy samples can efficiently contribute to achieving sufficient training samples;
(4) {\it w/o denoising} is worse than our framework, demonstrating that it is necessary to train the AHC in a noise reduction manner, e.g., fine-tuning in ours.


\begin{figure*}[htb]
\center
\subfigure[PEMS03 dataset, $P$-12/$Q$-12]{
\begin{minipage}[c]{0.4\linewidth} 
\centering
\includegraphics[width=\linewidth]{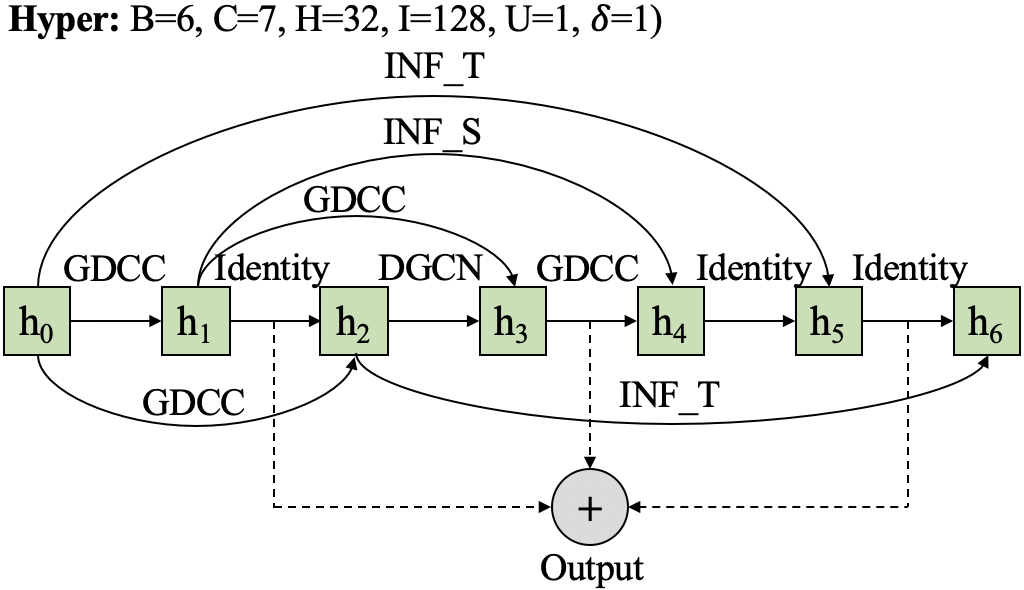}
\label{fig: search03}
\end{minipage}
}
\subfigure[PEMS08 dataset, $P$-48/$Q$-48]{
\begin{minipage}[c]{0.4\linewidth} 
\centering
\includegraphics[width=\linewidth]{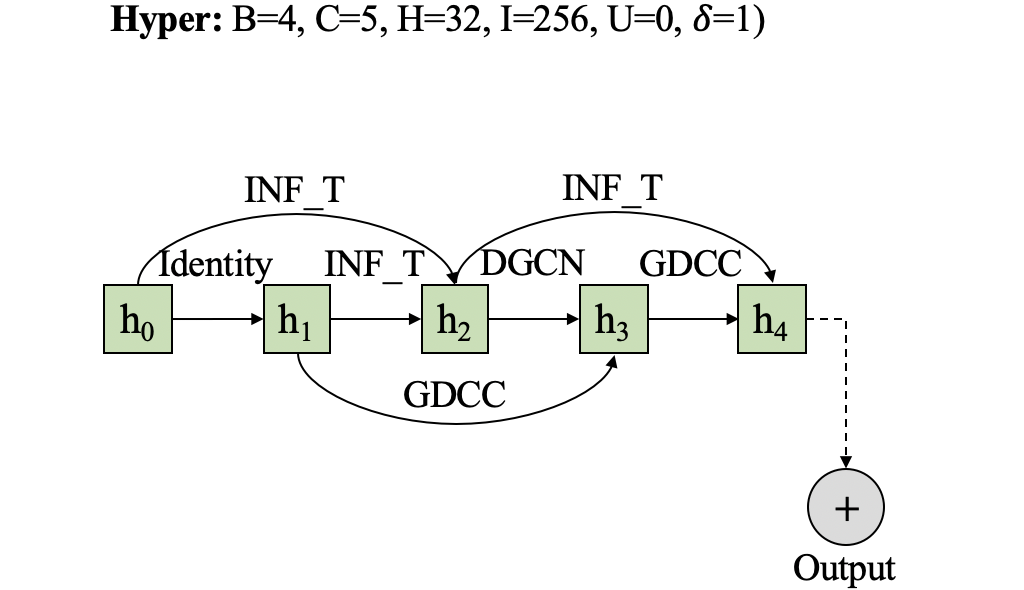}
\label{fig: search08}
\end{minipage}
}
\caption{Case Study of Searched ST-blocks on {Different} Target Datasets and $P$/$Q$ Settings.}
\label{fig: case}
\end{figure*}

\subsubsection{Efficiency Study}
We compare the search time of our framework with that of existing automated CTS forecasting frameworks in Table~\ref{tab: search_time}. 
For our framework, we consider the transfer setting and count the search time on target datasets.
This is because training AHC on the source dataset is usually a one-time job, so the corresponding search cost is not important.

%
It is obvious that our framework requires significantly less search time than AutoSTG and AutoCTS on the four datasets, which demonstrates the efficiency of our framework.

\begin{table}[htbp]
\small
    \centering
    \caption{Search Time Comparison in GPU hours for $P$-48/$Q$-48 Multi-step Forecasting.}
    \begin{tabular}{ccccc}
        \toprule  
        {Target Dataset}&AutoSTG&{AutoCTS}&{Ours-1}&Ours-2 \cr
    \midrule  
    PEMS03&584.3&{892.6}&190.4&\textbf{184.5} \cr
    PEMS04&316.6&{538.2}&\textbf{102.1}&- \cr
    PEMS07&1153.9&{2382.2}&447.7&\textbf{429.8} \cr
    PEMS08&261.4&{473.8}&-&\textbf{53.5} \cr
    \bottomrule
    \end{tabular}
    \label{tab: search_time}
\end{table}


\subsubsection{Impact of $L_1$, $L_2$, $z_1$, and $z_2$}\label{ssec: L1L2impact}


In the pre-training phase on the source dataset, we collect $L_1$ pairs of $(ah, R^{\prime}(ah))$ and $L_2$ pairs of $(ah, R(ah))$ to generate noisy and clean training samples for the AHC, respectively.
In the transfer phase, we use $z_1$ pairs of $(ah, R^{\prime}(ah))$ and $z_2$ pairs of $(ah, R(ah))$ to generate noisy and clean training samples to finetune the AHC.
The number of $L_1$, $L_2$, $z_1$, and $z_2$ affects the accuracy of the AHC.
It is expected that more clean and noisy samples lead to a more accurate AHC and thus better search results. 
We evaluate the impact of $L_1$, $L_2$, $z_1$, and $z_2$ on the PEMS08 dataset under the $P$-48/$Q$-48 setting.
{
To remove the bias caused by the sampling procedure, for each ($L_1$, $L_2$) or ($z_1$, $z_2$) setting, we train and test our framework five times using different $(ah, R^{\prime}(ah))$ or $(ah, R(ah))$ pairs chosen with five different random seeds, and we report the results in the format ``mean$\pm$standard deviation''.
}

\begin{table}[!htbp]
\small
    \centering
    \caption{Impact of $L_1$ and $L_2$  (mean$\pm$standard deviation). $\rho$ is the Spearman’s rank correlation.}
    \begin{tabular}{c|cccc}
    \toprule  
    ($L_1$, $L_2$)&MAE&RMSE&MAPE&$\rho$ \cr
    \hline
    (0, 150)&{18.79$\pm$0.11}&{29.16$\pm$0.13}&{14.34\%$\pm$0.16\%}&{0.78$\pm$0.02} \cr
    (1000, 0)&{18.42$\pm$0.09}&{28.69$\pm$0.16}&{14.22\%$\pm$0.17\%}&{0.80$\pm$0.02} \cr
    (1000, 100)&{18.28$\pm$0.08}&{28.51$\pm$0.12}&{13.54\%$\pm$0.13\%}&{0.83$\pm$0.01} \cr
    (1000, 150)&{18.10$\pm$0.06}&{28.29$\pm$0.09}&{13.35\%$\pm$0.10\%}&{0.85$\pm$0.01} \cr
    (2000, 100)&{17.83$\pm$0.09}&{27.92$\pm$0.15}&{13.08\%$\pm$0.14\%}&{0.87$\pm$0.02} \cr
    (2000, 150)&\textbf{17.71$\pm$0.07}&\textbf{27.90$\pm$0.12}&\textbf{12.56\%$\pm$0.12\%}&\textbf{0.89$\pm$0.01} \cr
    \bottomrule
    \end{tabular}
    \label{tab: L1}
\end{table}

\begin{table}[!htbp]
\small
    \centering
    \caption{Impact of $z_1$ and $z_2$ (mean$\pm$standard deviation). $\rho$ is the Spearman’s rank correlation.}
    \begin{tabular}{c|cccc}
    \toprule  
    $(z_1, z_2)$&MAE&RMSE&MAPE&$\rho$ \cr
    \hline
    (0, 0)&{18.76$\pm$0.05}&{28.92$\pm$0.08}&{14.27\%$\pm$0.09}&{0.79$\pm$0.00} \cr
    (100, 5)&{17.76$\pm$0.09}&{28.04$\pm$0.12}&\textbf{12.52\%$\pm$0.10\%}&\textbf{0.89$\pm$0.01} \cr
    (500, 10)&\textbf{17.64$\pm$0.09}&\textbf{27.81$\pm$0.11}&{12.90\%$\pm$0.15\%}&\textbf{0.89$\pm$0.01} \cr
    (1000, 20)&{17.85$\pm$0.10}&{27.86$\pm$0.12}&{12.82\%$\pm$0.13\%}&\textbf{0.89$\pm$0.01} \cr
    \bottomrule
    \end{tabular}
    \label{tab: L2}
\end{table}

From Table~\ref{tab: L1}, it is observed that with either increasing $L_1$ or $L_2$, the achieved performance increases.
In particular, when $L_1=2000$ and $L_2=150$, the trained AHC achieves a high $\rho$, which demonstrates that the trained AHC can produce a ranking close to the true ranking of arch-hypers by performing pairwise comparison.
%
%
Considering that the cost of a clean sample is about 20 times that of a noisy sample (100 training epochs vs 5 training epochs), and $\rho(1000, 0) > \rho(0, 150)$, the proposed proxy improves the sample efficiency by a factor of about 3. 

%
From Table~\ref{tab: L2} we can observe that:
(1) $(z_1=0, z_2=0)$ gets the worst performance and the worst AHC but the performance is still competitive compared to baselines (see Table~\ref{tab: main}). This demonstrates that good arch-hypers can be found even if we directly use the AHC trained on the source dataset, while further fine-tuning the AHC with a small number of noisy and clean samples leads to better performance;
(2) with $z_1$ larger than 100, and $z_2$ larger than 5, the performance saturates so we set $(z_1=100, z_2=5)$ as our default setting.

\subsubsection{Case Study}

We show two representative searched ST-blocks (arch-hypers) on different target datasets and settings in Figure~\ref{fig: case}. Figure~\ref{fig: search03} is searched on the PEMS03 dataset under the $P$-12/$Q$-12 setting, with PEMS08 as the source dataset; Figure~\ref{fig: search08} is searched on the PEMS08 dataset under the $P$-48/$Q$-48 setting, with PEMS04 as the source dataset.

We can see obvious differences between the two ST-blocks. In particular, the two ST-blocks have different hyperparameters: numbers of nodes $C$ (7 vs. 5), output modes $U$ (1 vs. 0), and output dimensions $I$ (128 vs. 256); for the architectures, they have totally different connections and operators between node pairs. These observations show that different datasets and settings prefer different arch-hypers, and our framework can successfully capture such preferences. 
{The reasons why different CTS datasets prefer different arch-hypers are as follows. 
First, different CTS datasets exhibit different temporal and spatial patterns, and different arch-hypers may be good at capturing different temporal and spatial patterns.
Second, different CTS datasets contain different numbers of time series.
}




\section{Related Work}
\label{sec: related}

\subsection{Manual Model Design. }

There are many manually designed models for forecasting correlated time series~\cite{DBLP:conf/sigmod/FaloutsosGJW19,DBLP:conf/cikm/ParkLBTJKKC20,DBLP:conf/www/Wang0WJWTJY20,lai2018modeling,shih2019temporal,DBLP:conf/ijcai/WuPLJZ19,bai2020adaptive,wu2020connecting,razvanicde2021}. The biggest difference between these works is that they design different ST-blocks, which determines their ability to capture temporal and spatial dependencies. 
Graph WaveNet~\cite{DBLP:conf/ijcai/WuPLJZ19} employs Diffusion GCN and gated dilated causal convolution to capture spatial and temporal dependencies, respectively, and sequentially stacks the two operators to build ST-blocks. AGCRN~\cite{bai2020adaptive} employs enhanced Chebyshev GCN and GRU to capture spatial and temporal dependencies, respectively. MTGNN~\cite{wu2020connecting} uses mix-hop graph convolution and dilated inception convolution to capture spatial and temporal dependencies, respectively.
We are inspired by these manually designed models when designing the architecture search space.


\subsection{Automated Model Design. }
Recently, several automated frameworks have been proposed for time series forecasting. 
%
AutoAI-TS~\cite{DBLP:conf/sigmod/ShahPVDCKSWBGGV21} provides an automated pipeline for time series forecasting, and it is able to select the most appropriate forecasting model from a set of existing forecasting models for a specific data set and forecasting setting. In contrast, the proposed SEARCH framework aims at automatically designing novel CTS forecasting models. 
AutoST~\cite{li2020autost} divides a city's time series into grids based on longitude and latitude, and forecasts CTS data using grid-based images containing values for each timestamp as input. This work is not applicable to our problem because the CTS dataset we use lacks latitude and longitude information, so grid-based images cannot be constructed.
AutoSTG~\cite{pan2019urban} and AutoCTS~\cite{wu2022autocts} are the two studies most relevant to this paper. AutoSTG designs a search space for CTS forecasting, and introduces meta-learning to learn the weights of architectures. AutoCTS focuses on designing a compact and effective search space for CTS forecasting, and achieves the best accuracy on multiple benchmark datasets. However, both methods are supernet-based, thus are not scalable and do not support joint search for architectures and hyperparameters.
There is also a large body of studies aim at automatically designing neural architectures for various tasks~\cite{liu2018darts,DBLP:conf/iclr/ZophL17,DBLP:conf/sigmod/ElMS20,DBLP:journals/pvldb/LiSZJLDZY00021,DBLP:conf/icde/LiCZZJCH21}.
BRP-NAS~\cite{dudziak2020brp} and CTNAS~\cite{chen2021contrastive} perform pairwise architecture comparisons to explore the search space to search for the best architecture. However, both methods do not support joint search for architectures and hyperparameters, and thus fail to address the first limitation.
AutoHAS~\cite{dong2020autohas} and FBNetV3~\cite{dai2021fbnetv3} are two studies that aim to automatically search for the best combination of architectures and hyperparameters. However, AutoHAS is also a supernet-based method, which has poor scalability and thus fails to address the second limitation. Furthermore, both methods are very inefficient. For example, FBNetV3 spends more than 10K GPU hours searching for the best arch-hyper, so it cannot solve the third limitation.

\section{Conclusion}
\label{sec: conclu}

We present a scalable and efficient joint search framework to automatically design high-performance ST-blocks for CTS forecasting. 
In particular, we first design a joint search space containing massive arch-hypers, each of which is a combination of an architecture and a hyperparameter setting. Next, we propose an AHC-based search strategy to explore the search space to find the best arch-hyper.
To improve the sample efficiency and accuracy of the AHC, we propose a proxy metric to generate noisy training samples and train the AHC in a noise reduction manner to alleviate the negative effects of wrongly labeled training samples. 
Besides, we propose a transfer method to further improve the efficiency of the framework, allowing us to transfer a pretrained AHC to unseen datasets to reduce the need for samples to train the AHC.
Comprehensive experiments on six commonly used correlated time series forecasting datasets demonstrate the effectiveness, scalability, and efficiency of the proposed framework.
As future work, it is of interest to design more accurate proxy metrics and denosing algorithms to further improve the efficiency of the proposed framework.
{In addition, the proposed AHC can be used in other domains such as computer vision and natural language processing, because the proposed joint search space, i.e., the arch-hyper graph, can also represent models used in other domains. We leave this as future work.
Next, the proposed framework only searches for the optimal arch-hyper based on the accuracy metric, while metrics such as latency and energy consumption may also need to be considered when deploying CTS forecasting models; thus, it is also of interest to design multi-objective CTS forecasting frameworks.
}

\bibliographystyle{ACM-Reference-Format}
\bibliography{acmart}


\end{document}